\documentclass[11pt,a4page]{article}
\usepackage[margin=1in]{geometry}
\usepackage{parskip}

\usepackage{graphicx} 
\usepackage{hyperref}
\usepackage[numbers]{natbib}
\usepackage{color}
\usepackage{times}

\usepackage{amssymb,amsmath,amsthm}
\usepackage{bbm}
\usepackage{subcaption}
\usepackage[normalem]{ulem}
\usepackage{booktabs}
\usepackage{multicol}
\usepackage{url}
\usepackage{thmtools}
\usepackage{comment}

\newtheorem{theorem}{Theorem}

\declaretheorem[name=Lemma]{lemma}

\newtheorem{remark}{Remark}

\usepackage[linesnumbered,ruled,vlined]{algorithm2e}

\usepackage[colorinlistoftodos, textwidth=4cm, shadow]{todonotes}

\newcommand{\db}{d^{\text{b}}}
\newcommand{\policyb}{\pi^{\text{b}}}
\newcommand{\Cclipped}{{C^\star_{\text{clipped}}}}

\newcommand{\KL}{\mathrm{KL}}

\newcommand{\Bern}{\mathrm{Bern}}

\newcommand{\Ham}{d_{\mathrm{Ham}}}
\newcommand{\PBoper}{\widehat{\mathcal{T}}_{\mathrm{pe}}}
\newcommand{\argmax}{\operatorname*{arg\,max}}

\renewcommand{\S}{\mathcal{S}}
\newcommand{\A}{\mathcal{A}}

\newcommand{\R}{\mathbb{R}}

\newcommand{\PP}{\mathbb{P}}
\newcommand{\E}{\mathbb{E}}

\newcommand{\NN}{\mathbb{N}}

\newcommand{\ERMVILCB}{ERM-VI-LCB}

\newcommand{\indOne}{\mathbbmss{1}}

\newcommand{\cA}{\mathcal{A}}
\newcommand{\cB}{\mathcal{B}}

\newcommand{\cD}{\mathcal{D}}

\newcommand{\cS}{\mathcal{S}}
\newcommand{\cT}{\mathcal{T}}

\newcommand{\cX}{\mathcal{X}}

\newcommand{\Vpe}{\widehat{V}_{\text{pe}}^{\pi^\star}}
\newcommand{\dmin}{d^\star_{\min}}

\newcommand{\BEAS}{\begin{eqnarray*}}
\newcommand{\EEAS}{\end{eqnarray*}}
\newcommand{\BEA}{\begin{eqnarray}}
\newcommand{\EEA}{\end{eqnarray}}
\newcommand{\BEQ}{\begin{equation}}
\newcommand{\EEQ}{\end{equation}}
\newcommand{\BIT}{\begin{itemize}}
\newcommand{\EIT}{\end{itemize}}
\newcommand{\BNUM}{\begin{enumerate}}
\newcommand{\ENUM}{\end{enumerate}}
\newcommand{\beq}{\begin{equation}}
\newcommand{\eeq}{\end{equation}}
\newcommand{\beqa}{\begin{eqnarray}}
\newcommand{\eeqa}{\end{eqnarray}}
\newcommand{\beqan}{\begin{eqnarray*}}
\newcommand{\eeqan}{\end{eqnarray*}}
\newcommand{\bealn}{\begin{align*}}
\newcommand{\eealn}{\end{align*}}

\newcommand{\el}{\end{flushleft}}
\newcommand{\bl}{\begin{flushleft}}

\renewcommand{\phi}{\varphi}
\renewcommand{\epsilon}{\varepsilon}

\def\NN{{\mathbb N}}

\def\PP{{\mathbb P}}
\def\RR{{\mathbb R}}

\title{Provably Efficient Offline Discounted Reinforcement Learning \\under Recursive Entropic Risk}

\author{Amirparsa Bahrami\footnote{School of Electrical Engineering, Sharif University of Technology. Email: {\tt\small amirparsa2828@gmail.com}.}
\and Oliver Mortensen\footnote{Department of Computer Science, University of Copenhagen. Email: {\tt\small olmo@di.ku.dk}.}
        \and Mohammad Sadegh Talebi\footnote{Department of Computer Science, University of Copenhagen. Email: {\tt\small sadegh.talebi@di.ku.dk}.}}

\date{\today}

\begin{document}

\maketitle

\begin{abstract}
We study risk-sensitive reinforcement learning in finite discounted infinite-horizon Markov Decision Processes with recursive entropic risk measures (ERM), parameterized by $\beta\neq 0$. We consider both risk-averse ($\beta>0$)  and risk-seeking ($\beta<0$) settings.  
We consider the offline RL setting, where the agent is provided with a fixed dataset and wishes to learn a near-optimal policy using this dataset.  
We introduce a model-based algorithm, called ERM Value Iteration Lower Confidence Bound (\ERMVILCB) and derive a PAC-type bound on its sample complexity. We also establish a corresponding lower bound, demonstrating that exponential dependence on $|\beta|/(1-\gamma)$ asserted by the upper bound is unavoidable in the worst case. The presented bounds provide the first rigorous sample complexity guarantees for offline RL under recursive ERM across both risk-averse and risk-seeking regimes. 
\end{abstract}

\section{Introduction}

The classical setting of Reinforcement learning (RL) is concerned with a \emph{risk-neutral} objective, where the underlying objective is to maximize the expected return, namely, the expected (possibly discounted) sum
of rewards \citep{sutton1998reinforcement}. 
However, this is arguably inadequate for many high-stakes
application domains, such as healthcare
\citep{ernst2006clinical}, operations research
\citep{delage2010percentile}, and transportation
\citep{kamran2020risk}. Such applications must account for the variability of returns and the associated risks. One principled approach to addressing this limitation is to optimize an objective
defined through a risk measure. Notable risk measures considered in the literature include
mean-variance \citep{li2000optimal}, value-at-risk (VaR)
\citep{dempster2002risk}, conditional VaR (CVaR)
\citep{shapiro2021LNs}, entropic risk measure (ERM)
\citep{howard1972risk}, and entropic VaR (EVaR)
\citep{ahmadi2012entropic}.
Among these, CVaR and ERM have received considerable
attention for modeling risk sensitivity in MDPs
\citep{chow2014algorithms,bisi2022risk,
bauerle_2011_markov,howard1972risk,borkar2002risk,
hau2023entropic,fei2020risk}.

In many of the application domains that motivate
risk-sensitive RL, active exploration of the environment
may also be costly, impractical, or undesirable due to
the potential consequences of exploratory decisions.
Instead, the agent may have access to a dataset collected
from past interactions with the environment and is tasked
with learning a near-optimal policy using this dataset,
without further exploration. This learning paradigm,
known as \emph{offline RL}, has received considerable
attention in recent years
\citep{rashidinejad2021bridging,jin2021pessimism,
li2024settling}.
A central challenge in offline RL arises from the
potential mismatch between the sampling distribution
underlying the dataset and the occupancy distribution
induced by an optimal policy. In the risk-sensitive
setting, this challenge is further compounded by
the nonlinear dependence of the objective on the
underlying transition probabilities.

In this paper, we study risk-sensitive offline RL in
discounted MDPs, where the agent's objective is defined
using the entropic risk measure (ERM), \emph{recursively}
applied to the reward-to-go at every decision-making step.
This is also referred to as the iterated, nested, or
dynamic formulation in various communities
\citep{asienkiewicz2017note,bauerle_markov_2022,
deng2025near,mortensen2026recursive}.
This formulation stands in contrast to the
\emph{non-recursive} (also called non-iterated or static)
formulation, where the risk measure is directly applied
to the total return (e.g.,
$\sum_{t=0}^\infty \gamma^t r_t$ in the discounted case)
\citep{borkar2002q,borkar2002risk,hau2023dynamic}.
The two formulations are not directly comparable, and
each may be preferred depending on the underlying
application; we refer to, e.g.,
\cite{deng2025near,mortensen2026recursive}
for further discussion.

While the sample complexity of risk-neutral offline RL
in tabular discounted MDPs is now well understood
\citep{rashidinejad2021bridging,li2024settling},
our understanding of its risk-sensitive counterpart
remains comparatively limited.

Recent work has established sample complexity guarantees
for learning under recursive ERM in discounted MDPs
in the generative setting
\citep{mortensen2026recursive,bahrami2026near},
as well as for learning under ERM in finite-horizon MDPs
in the forward interaction setting \citep{essakine2026tight}. 
These results, however, are obtained in the generative
setting, where the agent can obtain samples from the
MDP at arbitrary state-action pairs, and hence do not
account for the challenges associated with limited data
coverage in offline RL.
In the offline setting, existing sample complexity
guarantees for recursive ERM in discounted MDPs
\citep{zhang2024soft} are restricted to the risk-averse
regime and rely on stronger notions of concentrability.
This motivates the study of the fundamental sample
complexity of offline RL under recursive ERM and the
development of provably sample-efficient algorithms
under milder data coverage conditions.

\subsection{Main Contributions and Paper Organization}

We consider risk-sensitive offline RL in tabular discounted MDPs under 
recursive ERM under both the risk-averse and risk-seeking regimes. 
Learning performance is assessed in terms of sample complexity, defined as the total number $N$ of samples required, for given $(\varepsilon,\delta)$, to obtain an $\varepsilon$-optimal policy with probability exceeding $1-\delta$.

We make the following contributions. We develop a model-based algorithm, called \ERMVILCB, and establish PAC-type bounds on its sample complexity. \ERMVILCB\ applies the value iteration method to an MDP defined by the plug-in estimator for the transition kernel combined with reward penalization and, both of which are built using the offline dataset (and hence, with no further exploration).  
We show that the sample complexity of \ERMVILCB\ (Theorem \ref{thm:sample_complexity_ERMVILCB}) scales as\footnote{Throughout the paper, $\widetilde{\mathcal O}$ and $\widetilde{\Omega}$ suppress poly-logarithmic factors in the relevant problem parameters.} 
\begin{align*}
    \widetilde{O}\bigg(\frac{S\Cclipped}{\dmin(1-\gamma)^2\beta^2\varepsilon^2}\Big(e^{\beta/(1-\gamma)} + e^{-\beta/(1-\gamma)} - 2\Big)
    + 
    \frac{S\Cclipped }{\dmin(1-\gamma)|\beta| \varepsilon}\Big(e^{|\beta|/(1-\gamma)} -1\Big)   \bigg), 
\end{align*}
%
This bound holds for any discounted MDP with $S$ states, $A$ actions, discount factor $\gamma$, and risk parameter $\beta$, with $\beta>0$ (respectively, $\beta<0$) corresponding to a risk-averse (respectively, risk-seeking) agent; see Section \ref{Section:Background} for a precise definition.  Furthermore, $\Cclipped$ denotes the single-policy concentrability coefficient and $\dmin$ denotes the minimal occupancy distribution under the optimal policy over the state-space of the MDP. Both notions of occupancy distribution and concentrability coefficient are defined in a similar fashion as the risk-neutral setting (e.g., \cite{rashidinejad2021bridging,li2024settling}). 
A key feature of the presented sample complexity is the exponential dependence on the effective horizon $1/(1-\gamma)$, which is absent in the conventional risk-neutral setting, wherein $\beta=0$.  

We further establish a worst-case lower bound on the sample complexity of offline RL under recursive ERM. Specifically, we show in Theorem \ref{thm:LB_ERM} that at least 
\begin{align*}
\Omega \bigg(\frac{S\Cclipped}{(1-\gamma)\beta^2\varepsilon^2}\Big(e^{\beta/(1-\gamma)} + e^{-\beta/(1-\gamma)} - 2\Big)\bigg)
\,.
\end{align*}
samples are required to learn a $\varepsilon$-optimal policy with high probability.

The lower bound demonstrates that the exponential dependence on $|\beta|/(1\!-\!\gamma)$  in sample complexity upper bounds is unavoidable in the worst case. Notably, it verifies that the upper bound has the right form of exponential dependence on $|\beta|/(1\!-\!\gamma)$ in the interesting regime of small $\varepsilon$. The lower bound further establishes that the sample complexity bound for \ERMVILCB\ has an optimal dependence 
$S$, $\Cclipped$, and $\frac{1}{\varepsilon}$.

To the best of our knowledge, these results constitute the first upper and lower bounds on the sample complexity of recursive ERM in discounted MDPs in the offline RL setting. 

The remainder of this paper is organized as follows. Section \ref{Section:RelatedWork} reviews related work. Section \ref{Section:Background} introduces the necessary background and formal problem setup. Section  \ref{sec:ERMVILCB} presents the \ERMVILCB\ algorithm and its sample complexity bound, 
with proofs deferred to  Section \ref{Section:sample_complexity_analysis}. The lower bound is  presented in Section \ref{sec:lower-bound}. Finally, Section \ref{Section:Conclusion} concludes with a discussion and directions for future research. Omitted proofs are provided in the appendix.

\section{Related Work}
\label{Section:RelatedWork}

\subsection{Risk-neutral Offline RL}
There is a rich and growing literature on provably efficient offline RL in MDPs, covering a variety of settings; see, e.g.,
\citep{chen2019information,rashidinejad2021bridging,jin2021pessimism,ren2021nearly,
zhan2022offline,uehara2022pessimistic,uehara2022representation,li2024settling}.
A central challenge in offline RL is to account for the distribution mismatch between the offline dataset and the state-action visitation distribution induced by an optimal policy. A prominent approach to addressing this challenge
is based on the \emph{pessimism principle} \citep{rashidinejad2021bridging,jin2021pessimism},
which seeks to account for the uncertainty arising from insufficient data coverage when constructing a policy. While a large body of work focuses on tabular MDPs, a number of studies consider more general settings with function
approximation \citep{jin2021pessimism,zhan2022offline,
uehara2022representation}. Of particular relevance to our work is \cite{li2024settling}, who establish the minimax-optimal sample complexity of offline RL in tabular discounted MDPs, scaling as $\widetilde{\mathcal O}\big(
\frac{SC^\star_{\mathrm{clipped}}} {(1-\gamma)^3\varepsilon^2}\big)$,
where $C^\star_{\mathrm{clipped}}$ denotes the single-policy clipped concentrability coefficient. Their result demonstrates that a model-based algorithm combined with pessimistic planning can achieve optimal
sample complexity under this relatively mild coverage condition. Our algorithm follows a similar model-based pessimistic approach. However, existing sample complexity analyses in the risk-neutral setting crucially exploit
the additive structure of the return and the linearity of expectation, and hence do not directly extend to the recursive ERM objective considered in this paper.

\subsection{Risk-sensitive RL}
A substantial literature studies decision making under
risk measures in bandit and RL settings. In the bandit setting,  \cite{ahmadipour2025evar} establishes asymptotically
optimal sample complexity guarantees for best-arm identification under EVaR. 
Representative examples in RL include work on CVaR
\citep{deng2025near,du2023provably,chen2024provably,lam2022risk},
entropic risk
\citep{borkar2002risk,fei2020risk,fei2021risk,
fei2021exponential,moharrami2025policy,marthe2025efficient,
hau2023entropic},
and mean-variance risk \citep{huang2022achieving}.
Much of the literature on sample-efficient learning under
ERM focuses on episodic MDPs in the online setting, often
with regret as the performance criterion
\citep{fei2020risk,fei2021exponential,fei2021risk,
hu2023tighter,liang2024regret}.
More recently, \cite{essakine2026tight} studies policy learning
under ERM in finite-horizon MDPs with a forward interaction
model and establish tight sample complexity bounds.

For discounted MDPs under recursive ERM,
\cite{mortensen2026recursive} establishes PAC upper and
lower bounds on the sample complexity of both value and
policy learning in the generative setting. The analysis of \cite{mortensen2026recursive} is further refined by  \cite{bahrami2026near}, establishing improved sample complexity upper bounds that match the existing lower bounds in their exponential dependence on $|\beta|/(1-\gamma)$, thereby reducing the remaining gap to a polynomial factor in the effective
horizon. While these two works consider the same ERM as ours, both assume access
to a generative model of the MDP, and hence do not address the challenges arising from limited data coverage inherent to the offline RL setting.

Another line of work develops learning algorithms for entire families of risk measures, notably coherent risk measures \citep{petrik2012approximate,tamar2015policy,lam2022risk,zhao2024ra} and optimized certainty equivalents (OCEs) \citep{wang2025reductions,xu2023regret,
lee2025risk,mortensen2026OCE}. While ERM is not coherent, it belongs to the OCE family in the risk-averse regime (i.e., $\beta>0$). In particular, \cite{mortensen2026OCE} establishes PAC upper and lower bounds for learning under recursive OCEs in discounted MDPs in the generative setting.

\subsection{Risk-sensitive Offline RL}
Several works have investigated risk-sensitive RL in the offline setting; see, e.g., \citep{urpi2021risk,ma2021conservative,rigter2023one,zhang2024soft,
zhang2024pessimism,moghimi2026risk,wang2026OCEofflineRL}.
Among these, \cite{urpi2021risk,ma2021conservative} develop distributional RL algorithms for learning risk-sensitive policies from offline datasets.
\cite{rigter2023one} develop a model-based offline RL approach that accounts for both risk sensitivity and distribution shift through a common  risk-sensitive formulation. These works, however, do not establish PAC sample complexity guarantees for the problem considered in this paper.

A particularly relevant work is \cite{zhang2024soft}, who establish an equivalence between risk-sensitive MDPs and regularized robust MDPs.
Leveraging this equivalence, they develop a model-free offline RL algorithm, called robust fitted-Z iteration (RFZI), and establish sample complexity guarantees for discounted MDPs under recursive ERM in the risk-averse regime.
Their analysis accommodates function approximation but relies on a concentrability condition that requires coverage of distributions induced by arbitrary policies under a class of perturbed transition kernels.
In contrast, we consider tabular MDPs and develop a model-based pessimistic algorithm whose sample complexity is characterized in terms of a single-policy clipped concentrability coefficient.
Furthermore, our analysis covers both the risk-averse and risk-seeking regimes and establishes a corresponding sample complexity lower bound.

\cite{zhang2024pessimism} develop provably sample-efficient pessimistic offline RL algorithms for episodic linear MDPs under ERM applied to the cumulative return.
More recently, \cite{wang2026OCEofflineRL} studies offline RL in episodic MDPs under both recursive and non-recursive OCE objectives, thereby encompassing ERM in the risk-averse regime. They establish finite-sample upper and lower bounds
under single-policy concentrability in tabular MDPs and also consider extensions to linear function approximation. Both works focus on finite-horizon MDPs, and their  analyses exploit the finite-horizon structure.

Our work complements this literature by establishing sample complexity upper and lower bounds for offline RL under recursive ERM in tabular discounted infinite-horizon MDPs. In particular, our results characterize the
fundamental role of the single-policy clipped concentrability coefficient and the exponential dependence on $|\beta|/(1-\gamma)$ in the sample complexity of this problem.

\section{Background}
\label{Section:Background}

\paragraph{Notations.} For $n \in \NN$, let $[n] := \{1, \ldots, n\}$. $\mathbbmss 1\{\cdot\}$ denotes the indicator function. Given a set $\cX$, $\Delta(\cX)$ denotes the probability simplex over $\cX$. We use the convention that $\|\cdot \| := \|\cdot \|_\infty$ and explicitly use the subscript $\|\cdot \|_p$ when using $p$-norms for  $1\leq p<\infty$. The notation $L^\infty(\Omega, \mathcal{F},\mathbb{P})$ denotes the space of essentially bounded random variables on the probability space $(\Omega, \mathcal{F},\mathbb{P})$.

\subsection{Entropic Risk Measure}
For $\beta\neq0$, consider the exponential utility function defined as $u(t) = \frac{1}{\beta}(1-e^{-\beta t})$ for $t\in \mathbb{R}$. The utility $u$ is supposed to describe the preferences of some economic agent in the form of how much utility $u(t)$ she derives from some monetary quantity $t\in \mathbb{R}$. For any bounded random variable $X\in L^\infty(\Omega, \mathcal{F},\mathbb{P})$, it is evident that the associated \emph{certainty equivalent} to $u$ is $u^{-1}(\E[u(X)]) = \frac{-1}{\beta}\log(\E[e^{-\beta X}])$. 
We thus define the functional $\rho : L^\infty(\Omega,\mathcal{F},\mathbb{P}) \rightarrow \mathbb{R}$ by\footnote{We note that there is a lack of consensus regarding the sign of $\beta$ in the definition of ERM. We follow this convention considering its widespread use in the actuarial literature \citep{asienkiewicz2017note}.}
\begin{align}
\label{eq:ERM_functional}
    \rho(X) :=\rho(X; \beta):=-\frac{1}{\beta}\log\big(\E[e^{-\beta X}]\big)\,.
\end{align}

It is clear that we recover the risk-neutral case for $\beta \rightarrow 0$: $\lim_{\beta\to 0} \rho(X) = \E[X]$. We note that $\rho$ admits the following properties: 
\begin{align}
    \label{monotonicity_Phi}
    &\rho(X) \leq \rho(Y), \quad \text{for any }X\leq Y,
    \\
    \label{consistency_Phi}
    &\rho(c) = c, \quad \text{for any }c\in \mathbb{R},
    \\
    \label{jensen_beta>0}
    & \rho(X) \leq \E[X], \quad \text{for }\beta>0,
    \\
    \label{jensen_beta<0}
    & \rho(X) \geq \E[X], \quad \text{for }\beta<0.
\end{align}
Using $\rho$ as a measure of the preference for different random variables, it follows directly from (\ref{consistency_Phi})-(\ref{jensen_beta<0}) that $\rho(X)\leq \rho(\E[X])$ for $\beta>0$ and that $\rho(X)\geq \rho(\E[X])$ for $\beta<0$. It further shows that $\beta>0$ is associated with \emph{risk-aversion}, while $\beta<0$ is associated with a \emph{risk-seeking} behavior. It is well-known that  ERM for $\beta>0$ belongs to the class Optimized Certainty Equivalents (OCE). We refer the reader to Appendix A in \citep{mortensen2026recursive} for a primer on risk measures.

\subsection{Discounted Markov Decision Processes with Entropic Risk}
We write the 6-tuple $M = (\S,\A,P, R,\gamma, \beta)$ to denote a finite, discounted infinite-horizon MDP, where $\S = \{1,2,\ldots,S\}$ is the finite state space of size $S:=|\S|$, $\A = \{1,2,\ldots,A\}$ is the finite action space of size $A:=|\A|$, $P:\S\times \A\rightarrow \Delta(\S)$ is the transition probability function, $R:\S\times \A\rightarrow [0,1]$ is the reward function, $\gamma \in (0,1)$ is the discount factor, and $\beta\neq 0$ is the risk-parameter. 
We may write $P_{s,a}(s')$ as shorthand notation for $P(s'|s,a)$ for any $(s,a)\in \cS\times \cA$. For simplicity of exposition, we consider a deterministic reward function, as is standard in the literature. 
The agent interacts with the MDP $M$ as follows. At the beginning of the process, $M$ is in some initial state $s_0\in \S$ sampled from an initial distribution $\nu\in \Delta(\cS)$. At each time $t\ge 0$, the agent is in state $s_t\in \S$ and decides on an action $a_t\in \A$ according to some rule. The MDP generates a reward $r_t:=R(s_t,a_t)$ and a next-state $s_{t+1}\sim P(\cdot|s_t,a_t)$. The MDP moves to $s_{t+1}$ when the next time slot begins, and this process continues ad infinitum. This process yields a growing sequence $(s_t,a_t,r_t)_{t\ge 0}$. The agent's goal is to maximize an objective function, as a function of the collected rewards $(r_t)_{t\ge 0}$, which depends on both $\gamma$ and $\beta$. 

Let $v\in \mathbb{R}^S$ and $\pi:\S\rightarrow \A$ be a stationary deterministic policy. Inspired by the definition of ERM functional in Eq.~(\ref{eq:ERM_functional}), We  define $\rho_{s,a}:\mathbb{R}^S\rightarrow \mathbb{R}$ as 
\begin{align}
    \rho_{s,a}(v) = -\frac{1}{\beta}\log\Big(\E_{s'\sim P_{s,a}}[e^{-\beta v(s')}] \Big).
\end{align}
The agent's goal is to maximize a risk-sensitive objective of the collected rewards $(r_t)_{t\ge 0}$, depending on both $\gamma$ and $\rho$. In the classical setting,  the value of a policy $\pi$ is defined as discounted sum of rewards collected under $\pi$. 
Under recursive ERM, the objective is defined through the iteration of $\rho$ \citep{bauerle_markov_2024}. The value function $V^\pi$ and the state-action value function (or Q-value) $Q^\pi$ of a policy $\pi$ are defined informally as 
\begin{align*}
  V^\pi(s_0) & =r_0+\gamma\rho_{s_0,\pi(s_0)}\Big(r_1+\gamma\rho_{s_1,\pi(s_1)}\big(r_2 + \gamma \rho_{s_2,\pi(s_2)}(\ldots)\big)\Big)\,,
    \\
  Q^\pi(s_0,a) & =R(s_0,a)+\gamma\rho_{s_0,a}\Big(r_1+\gamma\rho_{s_1,\pi(s_1)}\Big(r_2 +\gamma \rho_{s_2,\pi(s_2)}(\ldots)\Big)\Big)\,,\quad \forall a\in \A,
\end{align*}
with $r_t:=R(s_t,\pi(s_t))$. For all $(s,a)$, let $V^\star(s):=\sup_{\pi}V^\pi(s)$ and $Q^\star(s,a):=\sup_\pi Q^\pi(s,a)$ denote the optimal value (at state $s$) and Q-value (at state-action $(s,a)$), respectively, where $\sup$ is taken over the set of all possible policies (potentially randomized and history-dependent). Any policy $\pi^\star$ satisfying $V^{\pi^\star}(s)=V^\star(s)$ for all $s\in \S$ is called optimal, and any policy $\pi$ obeying $V^\pi(s)\ge V^\star(s) - \varepsilon$ for a given $\varepsilon>0$ is called $\varepsilon$-optimal. 
%
As established in \cite{bauerle_markov_2024}, under a stationary $\pi$, 
\begin{align*}
    V^\pi(s)& =R(s,\pi(s))+\gamma \rho_{s,\pi(s)}(V^\pi), \quad &&\forall s\in \S.
    \\
    Q^\pi(s,a)& =\!R(s,a)+\gamma\rho_{s,a}\big(Q^\pi(\cdot, \pi(\cdot))\big), \quad &&\forall (s,a)\in \cS\times \cA.
\end{align*}
Furthermore, there exists a stationary deterministic optimal policy $\pi^\star:\cS\to\cA$ that achieves $V^\star(s)$ for all states $s$ simultaneously, and the optimal value functions satisfy the Bellman optimality equation:   
\begin{align*}
    V^\star(s)& =\max_{a\in \A}\big(R(s,a)+\gamma \rho_{s,a}(V^\star)\big), \quad &&\forall s\in \S,
   \\
    Q^\star(s,a)& =\!R(s,a)+\gamma\rho_{s,a}\big(\max_{a'\in \A}Q^\star(\cdot,a')\big), \quad &&\forall (s,a)\in \cS\times \cA.
\end{align*}
We introduce the Bellman operators $\mathcal{T}^\pi,\mathcal{T}:\mathbb{R}^{S}\rightarrow \mathbb{R}^{S}$, which for $v:\cS\rightarrow \mathbb{R}$ are defined as 
\begin{align}
\label{eq:V_Bellman_operator}
    (\mathcal{T}^{\pi}v)(s) &:=R(s,\pi(s))+\gamma\rho_{s,\pi(s)}\big(v(\cdot)\big), \\ 
    (\mathcal T v)(s) & :=\max_{s\in \cA}\Big(R(s,a)+\gamma\rho_{s,a}\big(v(\cdot)\big)\Big). 
\end{align}
Then, the Bellman equations for $V^\star$ and $V^\pi$ can  be written as $V^\star = \cT V^\star$ and $V^\pi = \cT^\pi V^\pi$. The operators $\mathcal{T}$ and $\mathcal{T}^\pi$ are $\gamma$-contractions with respect to the max-norm (see, \citep[Lemma~1]{mortensen2026recursive}). 

Finally, for the initial state distribution $\nu\in \Delta(\cS)$ and policy $\pi$, we introduce
$$
V^\pi(\nu) = \sum_{s\in \cS} \nu(s)V^\pi(s).
$$

\paragraph{Discounted Occupancy Distributions.}The discounted occupancy distribution under policy $\pi$ and state distribution $\nu$ is a function $d^\pi(\cdot;\nu): \cS\times\cA \to [0, 1]$ defined as: 
$$
d^\pi(s, a; \nu) := (1 - \gamma) \sum_{t = 0}^\infty \gamma^t \PP(s_t = s, a_t = a | s_0 \sim \nu; \pi), \quad \forall (s,a)\in \cS\times \cA.
$$
Slightly overloading notation, we introduce  
$d^\pi(s; \nu) := (1 - \gamma) \sum_{t = 0}^\infty \gamma^t \PP (s_t = s | s_0 \sim \nu; \pi)$ for all $s\in\cS$. 
Of particular interest will be occupancy distributions under the optimal policy: we introduce, for all $(s,a)$,
$$
d^\star(s) := d^{\pi^\star}(s; \nu), \qquad d^\star(s, a) := d^{\pi^\star}(s, a; \nu) = d^\star(s) \indOne\{a = \pi^\star(s)\}\,.
$$
Another relevant policy in the sequel will be the behavior policy $\policyb$ ---to be introduced in the following--- for which we define:
$
\db(s,a) := d^{\policyb}(s,a; \nu)$ for all  $(s,a)\in \cS\times \cA$.

\subsection{Offline RL under Recursive ERM}
The problem of offline RL in discounted MDPs under recursive ERM is formalized as follows. 

Assume that we have access to a batch dataset $\cD=\{(s_i,a_i,s'_i)\}_{i\in[N]}$ collected through interacting with an unknown (but fixed) MDP $M$ 
using a {\it behavior} policy $\policyb$. The policy $\policyb$ might be unknown to the agent. Similarly to most  literature on offline RL \citep{rashidinejad2021bridging,li2024settling}, we consider the \emph{independent sampling} setting, where the $N$ samples in $\cD$ are generated independently. More precisely, recalling that $\db\in \Delta(\cS\times \cA)$ denotes the occupancy distribution of $\policyb$, we have that: 
$$
\forall i\in [N]: \qquad (s_i, a_i) \stackrel{\text{ind.}}{\sim} \db \quad \hbox{and} \quad  s'_i\stackrel{\text{ind.}}{\sim} P(\cdot | s_i, a_i).
$$
In offline RL, the goal is to compute a near-optimal policy $\widehat\pi$ using the dataset $\cD$ (and without further exploration).
More precisely, for a pre-specified accuracy $\varepsilon\in (0,1/(1-\gamma)]$, we aim to find an $\varepsilon$-optimal policy $\widehat\pi$ satisfying
$V^\star(\nu) - V^\pi(\nu) \le \varepsilon$  with high probability 
using the smallest dataset $\cD$ possible.

\paragraph{Concentrability Coefficients.}In order to capture the mismatch in occupancy measure between the optimal policy and the behavior policy, concentrability notions are used.  Consider a policy $\policyb$ in MDP $M$. Two well-known notions are the \emph{single-policy concentrability coefficient} \citep{rashidinejad2021bridging}, denoted by $C^\star$, and the \emph{single-policy clipped concentrability coefficient} \citep{li2024settling}, denoted by $\Cclipped$. They are defined as 
$$
C^\star := \max_{(s, a) \in \cS \times \cA} \frac{d^\star(s, a)}{\db(s, a)}, \quad\qquad 
\Cclipped := \max_{(s, a) \in \cS \times \cA} \frac{\min\{ d^\star (s, a), \frac{1}{S}\}}{\db(s, a)}\,.
$$
Note that $C^\star_{\text{clipped}} \leq C^\star$ by construction. The coefficients $C^\star$ and $\Cclipped$ are originally defined for risk-neutral MDPs. 


\section{The \ERMVILCB\ lgorithm}
\label{sec:ERMVILCB}

In this section, we introduce a model-based algorithm for offline RL under recursive ERM. The algorithm, called \ERMVILCB, incorporates lower confidence bounds in value estimation and used a value iteration method to an empirical MDP with perturbed reward function based on suitable penalty terms. \ERMVILCB\ implement the pessimism principle analogously to VI-LCB \citep{li2024settling}, and may be seen as extension of VI-LCB from risk-neutral finite MDPs to the case of MDPs with objectives defined based on recursive ERM. 

Let us first introduce some necessary notations. For any pair $(s,a)$, we write $N(s,a)$ to denote the number of visits to $(s,a)$ in the dataset $\cD$. Similarly, for any triplet $(s,a,s')$, we define $N(s,a,s')$ to denote the number of times a visit to $(s,a)$ yielded the next-state $s'$. More formally, 
$$
N(s,a,s'):=\sum_{i=1}^N \indOne\{s_i=s, a_i=a, s'_i=s'\}, \qquad N(s,a) = \sum_{s'\in \cS} N(s,a,s').
$$
Using these counters $\{N(s,a), (s,a)\in \cS\times \cA\}$ and $\{N(s,a,s'), (s,a,s')\in \cS\times \cA\times \cS\}$, we build the empirical estimate $\widehat P$ for $P$  as follows:
    \begin{align}
        \label{eq:hatP}
        \forall (s, a, s') \in \cS \times \cA \times \cS, \qquad \widehat P(s'|s,a) = \frac{N(s, a, s')}{N(s, a)}\, .
    \end{align}

\paragraph{The pessimistic Bellman operator.} We need to introduce the following operator, called \emph{pessimistic ERM Bellman operator}, or pessimistic Bellman operator for short, which extends the pessimistic Bellman operator for risk-neutral setting \citep{li2024settling} to the setting of recursive ERM. The pessimistic Bellman operator $\PBoper:\R^{S}\rightarrow \R^S$ is defined such that for any $v\in \R^S$,   
\begin{align}\label{eq:pess_ERM_Bellman_operator}
        (\PBoper v)(s)
        :=
        \max\bigg\{0,
        \max_{a\in\cA}
        \bigg\{
        R(s,a)
        -\frac{\gamma}{\beta}
        \log\Big(\sum_x\widehat P(x| s,a)e^{-\beta v(x)}\Big)
        -b(s,a)
        \bigg\}
        \bigg\}.
\end{align}

This is defined to penalized the original Bellman operator in (\ref{eq:V_Bellman_operator}) with respect to the empirical transition function $\widehat P$. Here, $b(s,a)$ is a penalty term associated to the pair $(s,a)$ to enforce pessimism under uncertainty.. We introduce the precise form of $b(s,a)$ momentarily. We note that the variant for Q-value function may be defined straightforwardly. 

The following lemma shows that $\PBoper$ is a $\gamma$-contraction, thus resembling the classical Bellman operator for discounted MDPs.

\begin{lemma}[Contraction of Pessimistic Bellman Operator]
\label{lem:Tpe_contraction}
    The operator $\PBoper$ is a $\gamma$-contraction  w.r.t.~$\|\cdot\|$:
    $$
    \|\PBoper u - \PBoper v\| \le \gamma \|u - v\|,
    $$
    for all $u,v\in [0,1/(1-\gamma)]^S$. 
\end{lemma}

By the Banach fixed-point theorem, $\PBoper$ admits a unique fixed point, which we denote by $\widehat V^\star_{\rm pe}$: $\widehat V^\star_{\rm pe} = \PBoper \widehat V^\star_{\rm pe}$.

In general, the penalty function $b$ may depend on the (empirical) value functions. In this paper, we consider the following penalty function $b$:
{\small
\[
        b(s,a)
        :=
        \begin{cases}
        \dfrac1{1-\gamma}, & N(s,a)=0,\\[1.2em]
        \min\Bigg\{
        \dfrac1{1-\gamma},
        \dfrac{4\sqrt{2}}{|\beta|}
        \Bigg[
        \sqrt{
        \dfrac{
        \big(e^{|\beta|/(1-\gamma)}+e^{-|\beta|/(1-\gamma)}-2\big)\ell
        }{N(s,a)}}
        +
        \dfrac{
        \big(e^{|\beta|/(1-\gamma)}-1\big) \ell
        }{N(s,a)}
        \Bigg]
        +\dfrac{4}{N}
        \Bigg\}, & N(s,a)>0.
        \end{cases}
\]
}
with $\ell :=\ell(N,\delta) := \log\big(\frac{4SAN}{(1-\gamma)\delta}\big)$. The penalty term $b(s,a)$ is derived using a  Bernstein-type concentration which renders crucial to ensure that the final sample complexity bound has an near-optimal dependence on the term $|\beta|/(1-\gamma)$ in the worst-case.

\paragraph{The \ERMVILCB\ algorithm.} We are now ready to now introduce the \ERMVILCB\ algorithm. 


After constructing the empirical transition function $\widehat P$, the algorithm consists in doing a value iteration scheme based on $\PBoper$ for $k$ iterations, which results in an estimate value function $\widehat V:=\widehat V_k$. The algorithm outputs a policy $\widehat\pi$ which is the greedy policy in the penalized empirical MDP:
\[
\widehat \pi(s) \in \argmax_a \max\bigg\{0,R(s,a)-\frac{\gamma}{\beta}\log\Big({\sum_x\widehat{P}(x|s,a)e^{-\beta \widehat{V}_k(x)}\Big)-b(s,a)} \bigg\}\,.
\]
We set the number of iterations as 
$$
k \ge \frac{1}{1-\gamma}\log \frac{N}{1-\gamma}, 
$$
which ensures, using the contraction properties of $\PBoper$ (Lemma \ref{lem:Tpe_contraction}), that $\big\|\widehat V - \widehat V^\star_{\textrm{pe}}\big\|\le \frac{1}{N}$. 

\begin{algorithm}[H]
\label{Alg:QVI}
\SetAlgoLined
\SetKwInput{KwInput}{Input}                
\SetKwInput{KwOutput}{Output}              
\DontPrintSemicolon

  \KwInput{MDP parameters $S,A,R,\gamma,\beta$, dataset $\mathcal{D}$ and number of iterations $k$.}
  \KwOutput{Policy estimate $\widehat{\pi}$ of optimal policy $\pi^\star$ }
  Initialization: $\widehat V_0=0$ \;
  Compute $\widehat{P}(s'|s,a)$
  \;
\For{all $(s,a)\in \cS\times \cA$}{
  Compute penalty terms $b(s,a)$ according to either $b_W$ or $b_B$ } 
  \For{$j=1,\ldots,k$}{
  \For{all $s\in \cS$}{
    $\widehat V_{j}(s) = (\PBoper \widehat V_{j-1})(s)$  
  }
  }
  \KwRet policy $\widehat\pi(s) \in \argmax_a \max\{0,R(s,a)-\frac{\gamma}{\beta}\log({\sum_x\widehat{P}(x|s,a)e^{-\beta \widehat{V}_k(x)})-b(s,a)} \}$
  \caption{\ERMVILCB}
\end{algorithm}

\subsection{Performance  Guarantee}

We present theoretical performance guarantees for \ERMVILCB\ by analyzing its sample complexity. The following theorem offers a high-probability sample complexity bound for \ERMVILCB.  
Define $\dmin := \min_sd^\star(s)$. 

\begin{theorem}[{Sample Complexity of \ERMVILCB}]
    \label{thm:sample_complexity_ERMVILCB}
Let $\delta\in (0,1)$ and $\varepsilon\in (0,1/(1-\gamma)]$. Assume that $k\ge \frac{1}{1-\gamma}\log\big(\frac{N}{1-\gamma}\big)$. The policy $\widehat\pi$ returned by \ERMVILCB\ satisfies, 
$$
V^\star(\nu)-V^{\widehat\pi}(\nu)\le \varepsilon
$$
with probability at least $1-2\delta$, provided that the size $N$ of $\cD$ satisfies
    \begin{align*}
        N\ge c_1\frac{S\Cclipped }{\dmin (1-\gamma)^2\beta^2\varepsilon^2}\Big(e^{\beta/(1-\gamma)}+e^{-\beta/(1-\gamma)}-2\Big)\ell^2 +  c_2\frac{S\Cclipped }{\dmin(1-\gamma)|\beta| \varepsilon}\Big(e^{|\beta|/(1-\gamma)} -1\Big) \ell^2
    \end{align*}
with $c_1$ and $c_2$ being numerical constants and $\ell = \log\big(\frac{4SAN}{(1-\gamma)\delta}\big)$.
\end{theorem}

The presented sample complexity bound depends on $1/\dmin$
and is thus non-vacuous only when $d^\star(s)>0$ for all
$s\in\cS$. This amounts to requiring that every state
be reachable with positive probability under the optimal
policy from the initial state distribution $\nu$.
Removing the dependence on $1/\dmin$ remains an
interesting direction for future research.

\begin{remark}
    \cite{zhang2024soft} derive a sample complexity for offline RL under recursive ERM for $\beta>0$ for a model-free method in the function approximation setting. Their bound can be translated to the tabular setting by taking the function class to be that of all possible stationary policies (of size $A^S$). Their bound becomes   
    \begin{align*}
        \widetilde{O}\Big(\frac{SC^2e^{2\beta/(1-\gamma)}}{\varepsilon^2(1-\gamma)^4\beta^2}\Big)
    \end{align*}
    where $C$ is related to the \emph{all-policy concentrability coefficient}, which can be much larger than $\Cclipped$.
\end{remark}

\section{Sample Complexity Lower bound}
\label{sec:lower-bound}

In this section, we provide a sample complexity lower bound for offline RL under recursive ERM. 

\begin{theorem}[Lower Bound]
\label{thm:LB_ERM}
For any $(\gamma,S,A,C,\beta,\varepsilon)$ satisfying 
$\gamma\in[\frac34,1)$, $S\ge 8$, $A\ge 2$, $C\ge 1$, $\beta\neq 0$, and
$$
\varepsilon
\le
\frac{\gamma}{256|\beta|}
\left(
1-e^{-|\beta|/(1-\gamma)}\right),
$$
one can construct a family of discounted MDPs $\mathfrak M = \big\{M_\theta: \theta\in \{0,1\}^{\lceil S/4\rceil}\big\}$, 
an initial distribution $\nu$, and an offline sampling distribution $\db$, with associated clipped concentrability coefficients satisfying 
and the clipped concentrability coefficient satisfy
$\Cclipped(M_\theta,\db,\nu) =C$ for all $\theta$ 
such that 
\[
\inf_{\widehat\pi}
\sup_{\theta\in\{0,1\}^{\lceil S/4\rceil}}
\mathbb P_\theta
\left(
V^\star_{M_\theta}(\nu) - V^{\widehat\pi}_{M_\theta}(\nu) >
\varepsilon \right) \ge \frac18,
\]
provided that
\[
N \le \frac{1}{2^{20}}\frac{\gamma^2SC}{(1-\gamma)\beta^2\varepsilon^2}
\left[
\exp\left(\frac{\beta}{1-\gamma}\right) +
\exp\left(-\frac{\beta}{1-\gamma}\right) -2
\right].
\]
Here, $\PP_\theta$ denotes the probability when the underlying MDP is $M_\theta$, and the infimum is over all estimator $\widehat \pi$. 
\end{theorem}

The lower bound in Theorem \ref{thm:LB_ERM} tells us that the  sample complexity is at least scales as 
$$
\Omega \bigg(\frac{S\Cclipped}{(1-\gamma)\beta^2\varepsilon^2}
\left[
\exp\left(\frac{\beta}{1-\gamma}\right) +
\exp\left(-\frac{\beta}{1-\gamma}\right) -2
\right]\bigg)\,.
$$
Comparing this with the upper bound in Theorem \ref{thm:sample_complexity_ERMVILCB} indicates that \ERMVILCB\  enjoys a sample complexity that has as optimal dependence in $S,\Cclipped$, and $\frac{1}{\varepsilon}$. Furthermore, its dependence in the effective horizon has the same exponential form as the lower bound. However the upper bound of Theorem \ref{thm:sample_complexity_ERMVILCB} leaves a gap of order $\frac{1}{(1-\gamma)\dmin}$. Closing this gaps would be an interesting direction for future research. 

In the limit $\beta\rightarrow0$, the lower bound is of order $\Omega\big(\frac{S\Cclipped}{\varepsilon^2(1-\gamma)^3} \big)$ which is exactly the sample complexity in the risk-neutral case so in this sense our construction of the worst-case instance in the proof of Theorem \ref{thm:LB_ERM} is a strict generalization of the lower bound of \cite{li2024settling} to the full $\beta \in \R$ range. However, this construction comes at the expense of more conservative numerical constants.  

\begin{remark}
In the generative model setting with recursive ERM, \cite{mortensen2026recursive} reports a lower bound of order $\widetilde{\Omega}\big( \frac{SA}{\varepsilon^2\beta^2}\big(e^{|\beta|/(1-\gamma)}-3\big)\big)$, which becomes vacuous for small $\beta$. Just like the lower bound of this paper they use a construction where all decision states has two possible next state where one yields maximal reward and the other zero. The two confusing instances are thus characterized by their probabilities $p$ and $p+\alpha$ of transitioning to the good next state and by using the same parametrization of our lower bound it follows directly by their proof that the lower bound can be improved to $\widetilde{\Omega}\big(\frac{SA}{\varepsilon^2\beta^2}\big(e^{|\beta|/(1-\gamma)}+e^{-|\beta|/(1-\gamma)}-2\big)\big)$, which has the same scaling in all parameters and also converges to the meaningful bound of $\widetilde{\Omega}\big(\frac{SA}{\varepsilon^2(1-\gamma)^2} \big)$ in the risk-neutral $\beta\rightarrow 0$ limit. 
\end{remark}

\begin{remark}
    A lower bound for the offline setting with finite horizon and optimized certainty equivalents a lower bound is reported in \cite{wang2026OCEofflineRL}. Their bound is not directly comparable to ours as it is not a PAC-bound but a rate bound. Furthermore their bound is stated in terms of an optimizer that also depends on a hyperparameter choice so it is not very interpretable. 
\end{remark}

\section{Analysis: Proof of Theorem \ref{thm:sample_complexity_ERMVILCB}}
\label{Section:sample_complexity_analysis}
\subsection{Preliminaries}

First we prove some technical lemmas.

Let $E$ be the event that
\begin{align}\label{eq:event_E}   
\bigg|\frac{\gamma}{\beta}
\log
\frac{
\sum_x\widehat P(x| s,a)e^{-\beta v(x)}
}{
\sum_xP(x| s,a)e^{-\beta v(x)}
}
\bigg|
\le b(s,a).
\end{align}
holds for every $(s,a)\in \cS\times \cA$ and value function $v$ such that $\|v-\widehat V^\star_{\mathrm{pe}}\|\le 1/N$. 

The first lemma establishes that $E$ happens with high probability: 

\begin{lemma}[Event $E$]\label{lem:Bern-LOO}
We have that $\PP(E)\ge 1-\delta$. 
\end{lemma}

\begin{lemma}[{Pessimism}]
\label{lem:V_pessimism}
    On the event $E$, we have $V^{\widehat \pi}(s) \geq \widehat V(s)$ for all $s$.  
\end{lemma}

\begin{lemma}\label{lem:weighted-count-bound}
With probability at least $1-\delta$, 
one has, simultaneously, 
\begin{align*}
    \hbox{(i):}& \quad \sum_{(s,a):N(s,a)=0}d^\star(s,a)
        \le
        \frac{16S\Cclipped}{N}\log\left(\frac{SA}{\delta}\right),
        \\
\hbox{(ii):}&\quad 
        \sum_{s,a}\frac{d^\star(s,a)}{\max\{N(s,a),1\}}
        \le
        \frac{20S\Cclipped}{N}\log\left(\frac{SA}{\delta}\right).
\end{align*}
\end{lemma}

\begin{lemma}[Lemma B.1 in \cite{li2024settling}]
\label{lem:Nsa_Bernstein}
With probability at least $1 - \delta$, we have for all $(s,a)\in \cS\times \cA$:
$$
\max \Big\{ N(s,a), \frac{2}{3} \log \frac{NS}{\delta} \Big\} \geq \frac{N \db(s,a)}{12}.
$$
\end{lemma}

\subsection{Proof of Theorem \ref{thm:sample_complexity_ERMVILCB}}

\begin{proof}
Recall that $k = \frac{1}{1-\gamma}\log\frac{N}{1-\gamma}$. 
Since $\PBoper$ is a $\gamma$-contraction (Lemma \ref{lem:Tpe_contraction}) and the iterates stay in $[0,1/(1-\gamma)]^S$, the output $\widehat V:=\widehat V_k$ satisfies
\begin{align}\label{eq:Vhat_Vpe_diff}
\big\|\widehat V-\widehat V_{\mathrm{pe}}^\star\big\|
        \le
        \frac{\gamma^{k}}{1-\gamma}
        \le
        \frac{1}{N}.    
\end{align}

Assume that the event $E$ holds. Then, for any state $s$, 
\begin{align*}
    V^\star(s) - V^{\widehat\pi}(s) &\stackrel{\textrm{(i)}}{\le} V^\star(s) - \widehat V(s) 
    \stackrel{\textrm{(ii)}}{\le} V^\star(s) - \widehat V_{\mathrm{pe}}^\star(s) + \frac{1}{N} 
    \stackrel{\textrm{(iii)}}{\le} V^\star(s) - \widehat V_{\mathrm{pe}}^{\pi^\star}(s) + \frac{1}{N}, 
\end{align*}
where (i) follows from Lemma \ref{lem:V_pessimism}, (ii) from (\ref{eq:Vhat_Vpe_diff}), and (iii) uses the fact that $\widehat V^\star_{\mathrm{pe}} \ge \widehat V^{\pi^\star}_{\mathrm{pe}}$.  Hence, on the event $E$, 
\begin{align}\label{eq:value_decompos}
V^\star(\nu) - V^{\widehat\pi}(\nu) \le \sum_{s}\nu(s)\big(V^\star(s) - \widehat V^{\pi^\star}_{\mathrm{pe}}(s)\big) + \frac{1}{N}\,. 
\end{align}
Thus, it remains to control $V^\star(s) - \widehat V^{\pi^\star}_{\mathrm{pe}}(s)$ for all $s\in \cS$. 

Consider $s\in \cS$ and let $b^\star(s):=b(s,\pi^\star(s))$. By pessimistic Bellman equations,  we have 
\begin{align*}
    \big|V^\star(s)-\widehat{V}_{\mathrm{pe}}^{\pi^\star}(s)\big| & = \bigg|b^*(s)+\frac{\gamma}{\beta}\log\bigg(\frac{\sum_x\widehat{P}(x|s,\pi^\star(s))e^{-\beta \Vpe(x)}}{\sum_xP(x|s,\pi^\star(s))e^{-\beta V^\star(x)}} \bigg) \bigg|
    \\
    & \leq b^*(s) + \bigg|\frac{\gamma}{\beta}\log\bigg(\frac{\sum_x\widehat{P}(x|s,\pi^\star(s))e^{-\beta \Vpe(x)}}{\sum_xP(x|s,\pi^\star(s))e^{-\beta V^\star(x)}} \bigg) \bigg|
    \\
    & \leq b^*(s)+\gamma \|V^\star-\Vpe\|+\bigg|\frac{\gamma}{\beta}\log\bigg(\frac{\sum_x\widehat{P}(x|s,\pi^\star(s))e^{-\beta V^\star(x)}}{\sum_xP(x|s,\pi^\star(s))e^{-\beta V^\star(x)}} \bigg) \bigg|.
\end{align*}
Hence, 
\begin{align*}
    \|V^\star-\Vpe\| \leq \max_{s}\bigg\{ \frac{b^*(s)}{1-\gamma} +\frac{1}{1-\gamma}\bigg|\frac{\gamma}{\beta}\log\bigg(\frac{\sum_x\widehat{P}(x|s,\pi^\star(s))e^{-\beta V^\star(x)}}{\sum_xP(x|s,\pi^\star(s))e^{-\beta V^\star(x)}} \bigg) \bigg| \bigg\}
\end{align*}

and so by Lemma \ref{lem:bernstein}, we thus have that on the event $E$, 
\begin{align*}
    \big\|V^\star-\widehat{V}_{\mathrm{pe}}^{\pi^\star}\big\|\leq \frac{2}{1-\gamma}\max_{s} b^\star(s).
\end{align*}


Writing $\cS^+=\{s:N(s,\pi^\star(s))>0\}$, we have that
\begin{align*}
    \max_s b^\star(s) & \leq \sum_{s\in \cS\setminus\cS^+}b^\star(s) +\max_{s\in \cS^+}b^\star(s)
    \\
    & \leq \frac{1}{\dmin}\frac{1}{1-\gamma}\sum_{s\in \cS\setminus \cS^+}d^\star(s)+\max_{s\in \cS^+}b^\star(s) \\
    &\le \frac{16S\Cclipped\log(SA/\delta)}{d^\star_{\min}(1-\gamma)N} + \max_{s\in \cS^+}b^\star(s)\\
    &\le \frac{16S\Cclipped\ell}{d^\star_{\min}(1-\gamma)N} + \max_{s\in \cS^+}b^\star(s) \,,
\end{align*}
where in the second line we used that $b^\star(s) = 1/(1-\gamma)$ for $s\notin \cS^+$, in the third line we used Lemma \ref{lem:weighted-count-bound}, and in the last line we used $\ell\ge \log(SA/\delta)$.  

Define 
\[
h = \frac{e^{\beta/(1-\gamma)}+e^{-\beta/(1-\gamma)}-2}{\beta^2}\qquad \hbox{and} \qquad h' =\frac{e^{|\beta|/(1-\gamma)} -1}{|\beta|}.
\]
For the second term,  
we have
\begin{align*}
    \max_{s\in \cS^+}b^\star(s) & \leq 4\sqrt{2}\bigg[\sqrt{\frac{h \ell}{\min_{s\in \cS^+}N(s,\pi^\star(s))}}+\frac{h'\ell}{\min_{s\in \cS^+}N(s,\pi^\star(s))}\bigg] + \frac{2}{N}
    \\
    & \leq 4\sqrt{2}\bigg[\sqrt{\sum_{s\in \cS^+}\frac{h\ell}{N(s,\pi^\star(s))}}+\sum_{s\in \cS^+}\frac{h'\ell}{N(s,\pi^\star(s))}\bigg]+\frac{2}{N}
    \\
    & \leq 4\sqrt{2} \bigg[\sqrt{\frac{h\ell}{\dmin}\sum_{s\in \cS^+}\frac{d^\star(s)}{N(s,\pi^\star(s))}}+\frac{h' \ell}{\dmin}\sum_{s\in \cS^+}\frac{d^\star(s)}{N(s,\pi^\star(s))}\bigg] + \frac{2}{N}.
\end{align*}

By Lemma \ref{lem:weighted-count-bound} and using $\ell\ge \log(SA/\delta)$, we thus have with probability at least $1-\delta$ that 
\begin{align*}
\max_{s\in \cS^+}b^\star(s) & \leq 4\sqrt{2}\bigg[\sqrt{\frac{h\ell^2}{\dmin}\frac{20S\Cclipped}{N}}+\frac{h'\ell^2}{\dmin}\frac{20S\Cclipped}{N}\bigg]+\frac{2}{N}\, .
\end{align*}
Putting together, and using $\ell\ge \log(SA/\delta)$, we get 
\begin{align*}
    \big\|V^\star-\widehat{V}_{\mathrm{pe}}^{\pi^\star}\big\|&\leq \frac{32S\Cclipped\ell}{\dmin (1-\gamma)^2N} + \frac{8\sqrt{2}\ell}{1-\gamma}\sqrt{\frac{20S\Cclipped h}{\dmin N}}+\frac{160\sqrt{2}S\Cclipped\ell^2 h'}{\dmin(1-\gamma) N} + \frac{4}{(1-\gamma)N}
\end{align*}

Using (\ref{eq:value_decompos}) and after some elementary calculations, we have that if we require 
$$
N\ge c_1\frac{S\Cclipped h}{\dmin (1-\gamma)^2\varepsilon^2}\ell^2 +  c_2\frac{S\Cclipped h'}{\dmin(1-\gamma) \varepsilon} \ell^2
$$
for  $c_1\ge 10240$ and $c_2\ge 640$, then 
$V^\star(\nu)-V^{\widehat\pi}(\nu)\le \varepsilon$
probability at least $1-2\delta$. 
\end{proof}

\section{Conclusions}
\label{Section:Conclusion}

We studied offline RL in finite discounted MDPs, where the agent exhibits recursive risk preferences modeled via the entropic risk measure (ERM). We introduced a model-based algorithm, \ERMVILCB, and derived PAC-type bounds on its sample complexity. In addition, we established a corresponding sample complexity lower bound. The lower bound shows that an exponential dependence on the horizon $1/(1-\gamma)$ is unavoidable in the worst case, demonstrating that this setting is fundamentally harder than the risk-neutral offline RL. To our knowledge, these constitute the first upper and lower bounds for offline RL under recursive ERM. Closing the gaps between the lower and upper bound constitute an interesting direction for future work. 
Another promising direction is the development of model-free algorithms for this setting and the analysis of their statistical efficiency.

\bibliography{references}
\bibliographystyle{unsrt}

\appendix

\section{Proofs of the Upper Bound}

\subsection{Proof of Lemma \ref{lem:Tpe_contraction}}
\begin{proof}
We consider the case of $\beta>0$. (The case of $\beta<0$ admits a similar proof.)

Let $u,v\in [0,1/(1-\gamma)]^S$, and let $s$ be a state such that $\big\|\PBoper v- \PBoper u \big\| =\big|(\PBoper v)(s)- (\PBoper u)(s)\big|$. We assume without loss of generality that $\PBoper v(s)\geq \PBoper u(s)$.  This leaves us with three cases to check:

\paragraph{Case 1: $(\PBoper v)(s) = (\PBoper u)(s) = 0$.} This case holds trivially.

\paragraph{Case 2: $(\PBoper v)(s) > 0$ and $(\PBoper u)(s) > 0$.} Using that $\max_xf(x)-\max_xg(x)\leq \max_x(f(x)-g(x))$, we have 
\begin{align*}
    \big\|\PBoper v-\PBoper u\big\| & = \max_a\bigg\{R(s,a)-b(s,a)-\frac{\gamma}{\beta}\log\big(\sum_xP(x|s,a)e^{-\beta v(x)} \big)   \bigg\}
    \\& -\max_a\bigg\{R(s,a)-b(s,a)-\frac{\gamma}{\beta}\log\big(\sum_xP(x|s,a)e^{-\beta u(x)} \big)   \bigg\}
    \\
    & \leq \max_a\frac{\gamma}{\beta}\log\bigg(\frac{\sum_xP(x|s,a)e^{-\beta u(x)}}{\sum _xP(x|s,a)e^{-\beta v(x)}} \bigg)
    \\
    & = \max_a\frac{\gamma}{\beta}\log\bigg(\frac{\sum_xP(x|s,a)e^{-\beta u(x)}}{\sum _xP(x|s,a)e^{-\beta (v(x)-u(x))-\beta u(x)}}\bigg) 
    \\
    & \leq \max_a\frac{\gamma}{\beta}\log\bigg(\frac{\sum_xP(x|s,a)e^{-\beta u(x)}}{\sum _xP(x|s,a)e^{-\beta \|v(x)-u(x)\|-\beta u(x)}}\bigg) 
    \\
    & = \leq \max_a\frac{\gamma}{\beta}\log\bigg(\beta\|v-u\|\frac{\sum_xP(x|s,a)e^{-\beta u(x)}}{\sum _xP(x|s,a)e^{-\beta u(x)}}\bigg) 
    \\
    & \leq \gamma \|v-u\|.
\end{align*}

\paragraph{Case 3: $(\PBoper v)(s) > 0$ and $(\PBoper u)(s) = 0$.}
In this case it holds that 
\[
\max_a\bigg\{R(s,a)-b(s,a)-\frac{\gamma}{\beta}\log\big(\sum_xP(x|s,a)e^{-\beta u(x)} \big)   \bigg\}\leq 0.
\]
Hence, 
\begin{align*}
    \big\|\PBoper v-\PBoper u\big\| & = \max_a\bigg\{R(s,a)-b(s,a)-\frac{\gamma}{\beta}\log\big(\sum_xP(x|s,a)e^{-\beta v(x)} \big)   \bigg\}
    \\
    & \leq \max_a\bigg\{R(s,a)-b(s,a)-\frac{\gamma}{\beta}\log\big(\sum_xP(x|s,a)e^{-\beta v(x)} \big)   \bigg\}
    \\
    & - \max_a\bigg\{R(s,a)-b(s,a)-\frac{\gamma}{\beta}\log\big(\sum_xP(x|s,a)e^{-\beta u(x)} \big)   \bigg\}
    \\
    & \leq \gamma \|v-u\|.
\end{align*}
where we have skipped the lines that are similar to Case 2. 
\end{proof}

\subsection{Proof of Lemma \ref{lem:Bern-LOO}}

In this subsection we prove Lemma \ref{lem:Bern-LOO}. First, we give a lemma that will be used in the proof, which itself is proven at the end of this subsection): 

\begin{lemma}\label{lem:bernstein}
Fix a state-action pair $(s,a)$ and condition on $N(s,a)$.
Let $X_1,\ldots,X_{N(s,a)}$ be the next-state samples drawn from $P(\cdot|s,a)$.
Let $v\in[0,1/(1-\gamma)]^S$ be independent of these samples.
If $N(s,a)>0$, then with probability at least $1-\alpha$,
\[
\begin{aligned}
&\gamma
\left|
\frac1\beta
\log\frac{\sum_{x\in\cS}\widehat P(x|s,a)e^{-\beta v(x)}}{\sum_{x\in\cS}P(x| s,a)e^{-\beta v(x)}}
\right|\le
\frac{4\sqrt{2}}{|\beta|}
\Bigg[
\sqrt{
\frac{
\left[e^{\frac{|\beta|}{1-\gamma}}+e^{-\frac{|\beta|}{1-\gamma}}-2\right]\log(2/\alpha)
}{N(s,a)}}
+
\frac{
(e^{\frac{|\beta|}{1-\gamma}}-1)\log(2/\alpha)
}{N(s,a)}
\Bigg].
\end{aligned}
\]
If $N(s,a)=0$, then deterministically,
\[
\begin{aligned}
\gamma
\left|
\frac1\beta
\log\frac{\sum_{x\in\cS}\widehat P(x|s,a)e^{-\beta v(x)}}{\sum_{x\in\cS}P(x| s,a)e^{-\beta v(x)}}
\right|
\le
\frac1{1-\gamma}.
\end{aligned}
\]
\end{lemma}

The value function to which the event $E$ is applied in the proof of Theorem~\ref{thm:sample_complexity_ERMVILCB} is the finite value-iteration output $v=\widehat V$. Since $\widehat V$ is computed from the empirical transition kernel $\widehat P$, it generally depends on the same next-state samples that form $\widehat P(\cdot|s,a)$. Hence Lemma~\ref{lem:bernstein}, which requires independence from these
samples, cannot be applied directly. We remove this dependence by the leave-one-out construction of
\cite[Section~4.2]{agarwal_model-based_2020}, also used in
\cite{li2024settling}.

\begin{proof}
\noindent\textbf{Step 1: construction of auxiliary state-absorbing MDPs.}
For each
$s\in\cS$ and each
$u\in[0,1/(1-\gamma)]$, construct an auxiliary empirical MDP
$\widehat M_{s,u} = (\cS, \cA, \widehat P_{s,u}, R_{s,u}, \gamma)$, with 
\[
\begin{aligned}
\widehat P_{s,u}(\widetilde s\mid s,a)
&=
\indOne\{\widetilde s=s\},
&& \qquad
\text{for all }(\widetilde s,a)\in\cS\times\cA,
\\
\widehat P_{s,u}(\cdot\mid s',a)
&=
\widehat P(\cdot\mid s',a),
&& \qquad
\text{for all }(s',a)\in\cS\times\cA,\quad s'\neq s,
\\
R_{s,u}(s,a)
&=
(1-\gamma)u,
&& \qquad
\text{for all }a\in\cA,
\\
R_{s,u}(s',a)
&=
R(s',a),
&& \qquad
\text{for all }(s',a)\in\cS\times\cA,\quad s'\neq s.
\end{aligned}
\]
Thus, $\widehat M_{u,s}$ is identical to $\widehat M$ except at state $s$, and all randomness due to  empirical transition leaving state $s$ is
removed and $s$ is made absorbing with prescribed value $u$. 
Define \[
b_{s,u}(s',a)
=
\begin{cases}
0, & s'=s,\\
b(s',a), & s'\neq s.
\end{cases}
\]
For any $v\in[0,1/(1-\gamma)]^S$, define the auxiliary pessimistic Bellman operator 
\begin{align*}
(\widehat{\mathcal T}_{\mathrm{pe},s,u}v)(s')
&:=
\max\bigg\{0,\,
\max_{a\in\cA}
\bigg[
R_{s,u}(s',a)
-\frac{\gamma}{\beta}
\log\Big(
\sum_x\widehat P_{s,u}(x\mid s',a)e^{-\beta v(x)}
\Big)
-b_{s,u}(s',a)
\bigg]
\bigg\}.
\end{align*}
Let $\widehat V_{s,u}$ denote its fixed point. The operator is a $\gamma$-contraction, so the fixed point is unique by Banach's fixed-point theorem. 

{Condition} on the sampled state-action pairs, all counts $N(s',a')$, and hence all penalties, are deterministic. Conditional on these state-action pairs, the next-state samples are independent across observations and are distributed according to their corresponding transition rows. By construction, $\widehat P(\cdot|s,a)$ is not used anywhere in the auxiliary Bellman equations. Therefore, for every fixed $u$, $\widehat V_{s,u}$ is independent of the samples forming $\widehat P(\cdot|s,a)$ for every $a\in\cA$.

\noindent\textbf{Step 2: correspondence with the original empirical fixed points.}
Let $u^\star=\widehat V_{\rm pe}^\star(s)$. We have
\[
(1-\gamma)u^\star
-\frac{\gamma}{\beta}\log e^{-\beta u^\star}
=(1-\gamma)u^\star+\gamma u^\star
=u^\star.
\]
Thus, $\widehat V_{\rm pe}^\star$ satisfies the auxiliary Bellman equation at $s$. Since $\widehat M_{s,u^\star}$ is identical to $\widehat M$ at every state $s'\neq s$, $\widehat V_{\rm pe}^\star$ also satisfies the auxiliary Bellman equations there. Therefore, by uniqueness of the fixed point,
\[
\widehat V_{s,\widehat V_{\rm pe}^\star(s)}
=
\widehat V_{\rm pe}^\star.
\]

\noindent\textbf{Step 3: Construction of a finite cover.}
Let
$
u,u'
\in
[0,1/(1-\gamma)]$. For any fixed
$v\in[0,1/(1-\gamma)]^S$, $\widehat{\mathcal T}_{{\rm pe},s,u}$ and
$\widehat{\mathcal T}_{{\rm pe},s,u'}$ are identical at every
state $s'\neq s$, because changing $u$ only changes the prescribed
value of the absorbing state. At state $s$,
\[
(\widehat{\mathcal T}_{{\rm pe},s,u}v)(s)
=
(1-\gamma)u+\gamma v(s) \qquad \hbox{and}\qquad 
(\widehat{\mathcal T}_{{\rm pe},s,u'}v)(s)
=
(1-\gamma)u'+\gamma v(s),
\]
so that 
$\big\|
\widehat{\mathcal T}_{{\rm pe},s,u}v
-
\widehat{\mathcal T}_{{\rm pe},s,u'}v
\big\|
=
(1-\gamma)|u-u'|
$. 
Hence,  
\begin{align*}
\big\|\widehat V_{s,u}-\widehat V_{s,u'}\big\|
&\le
\big\|
\widehat{\mathcal T}_{{\rm pe},s,u}\widehat V_{s,u}
-
\widehat{\mathcal T}_{{\rm pe},s,u}\widehat V_{s,u'}
\big\|
+
\big\|
\widehat{\mathcal T}_{{\rm pe},s,u}\widehat V_{s,u'}
-
\widehat{\mathcal T}_{{\rm pe},s,u'}\widehat V_{s,u'}
\big\|
\\
&\le
\gamma
\big\|\widehat V_{s,u}-\widehat V_{s,u'}\big\|
+
(1-\gamma)|u-u'|, 
\end{align*}
Therefore
$\big\|\widehat V_{s,u}-\widehat V_{s,u'}\big\|
\le
|u-u'|$. 

Combining this with Step 2 gives $\big\|\widehat V_{\rm pe}^\star-\widehat V_{s,u}\big\|
\le
|\widehat V_{\rm pe}^\star(s)-u|$. For each state $s$, construct the grid
\[
\mathcal U_s
:=
\left\{
\frac{2j}{N}:
0\le j\le
\left\lfloor\frac{N}{2(1-\gamma)}\right\rfloor
\right\}
\cup
\left\{\frac1{1-\gamma}\right\}.
\]
Then, $|\mathcal U_s|
\le \frac{2N}{1-\gamma}$ and $\min_{u\in\mathcal U_s}|z-u|
\le \frac1N$, for every
$z\in[0,1/(1-\gamma)]$. 

\noindent\textbf{Step 4: concentration on the auxiliary values.}
For every fixed $(s,a,u)$ with $u\in\mathcal U_s$, the auxiliary value $\widehat V_{s,u}$ is independent of the samples forming $\widehat P(\cdot|s,a)$, which enables applying Lemma~\ref{lem:bernstein}. Taking
\[
\alpha=\frac{(1-\gamma)\delta}{2SAN}
\]
and using a union bound (and noting that the number of triples $(s,a,u)$ is at most $2SAN/(1-\gamma)$), it holds simultaneously for all $(s,a)$ with $N(s,a)>0$ and all $u\in\mathcal U_s$, with probability at least $1-\delta$, 
\begin{align}
\label{eq:loo-aux-bound}
\frac{\gamma}{|\beta|}
\left|
\log
\frac{
\sum_x\widehat P(x| s,a)
e^{-\beta\widehat V_{s,u}(x)}
}{
\sum_xP(x| s,a)
e^{-\beta\widehat V_{s,u}(x)}
}
\right| \le
\frac{4\sqrt{2}}{|\beta|}
\Bigg[
\sqrt{
\frac{
\left[
e^{\frac{|\beta|}{1-\gamma}}
+
e^{-\frac{|\beta|}{1-\gamma}}
-2
\right]\ell 
}{
N(s,a)
}
}
+
\frac{
\left(
e^{\frac{|\beta|}{1-\gamma}}-1
\right)\ell 
}{
N(s,a)
}
\Bigg].
\end{align}

For each state $s$, choose a grid point
$u_s\in\mathcal U_s$ satisfying $|\widehat V_{\rm pe}^\star(s)-u_s| \le
1/N$. 
Although $u_s$ depends on the data, this causes no difficulty because
\eqref{eq:loo-aux-bound} holds simultaneously for every
$u\in\mathcal U_s$.

By the results of Steps~2--3,
\[
\begin{aligned}
\big\|\widehat V_{\rm pe}^\star-\widehat V_{s,u_s}\big\|
&=
\big\|\widehat V_{s,\widehat V_{\rm pe}^\star(s)}-\widehat V_{s,u_s}\big\|
\le
|\widehat V_{\rm pe}^\star(s)-u_s|
\le\frac1N.
\end{aligned}
\]

We now relate the bound for $\widehat V_{s,u_s}$ to $\widehat V_{\rm pe}^\star$.
\[
\begin{aligned}
\log
\frac{
\sum_xP(x|s,a)e^{-\beta\widehat V_{\rm pe}^\star(x)}
}{
\sum_x\widehat P(x|s,a)e^{-\beta\widehat V_{\rm pe}^\star(x)}
}
&=
\log\sum_xP(x|s,a)e^{-\beta\widehat V_{\rm pe}^\star(x)}
-
\log\sum_xP(x|s,a)e^{-\beta\widehat V_{s,u_s}(x)}
\\
&\quad+
\log\sum_xP(x|s,a)e^{-\beta\widehat V_{s,u_s}(x)}
-
\log\sum_x\widehat P(x|s,a)e^{-\beta\widehat V_{s,u_s}(x)}
\\
&\quad+
\log\sum_x\widehat P(x|s,a)e^{-\beta\widehat V_{s,u_s}(x)}
-
\log\sum_x\widehat P(x|s,a)e^{-\beta\widehat V_{\rm pe}^\star(x)}.
\end{aligned}
\]
By the triangle inequality,
\[
\begin{aligned}
\left|
\log
\frac{
\sum_xP(x|s,a)e^{-\beta\widehat V_{\rm pe}^\star(x)}
}{
\sum_x\widehat P(x|s,a)e^{-\beta\widehat V_{\rm pe}^\star(x)}
}
\right|
&\le
\left|
\log
\frac{
\sum_xP(x|s,a)e^{-\beta\widehat V_{\rm pe}^\star(x)}
}{
\sum_xP(x|s,a)e^{-\beta\widehat V_{s,u_s}(x)}
}
\right|
\\
&\quad+
\left|
\log
\frac{
\sum_xP(x|s,a)e^{-\beta\widehat V_{s,u_s}(x)}
}{
\sum_x\widehat P(x|s,a)e^{-\beta\widehat V_{s,u_s}(x)}
}
\right|
+
\left|
\log
\frac{
\sum_x\widehat P(x|s,a)e^{-\beta\widehat V_{s,u_s}(x)}
}{
\sum_x\widehat P(x|s,a)e^{-\beta\widehat V_{\rm pe}^\star(x)}
}
\right|.
\end{aligned}
\]
The first term on the right-hand side is bounded as
\begin{align*}
\left|
\log
\frac{
\sum_xP(x|s,a)e^{-\beta\widehat V_{\rm pe}^\star(x)}
}{
\sum_xP(x|s,a)e^{-\beta\widehat V_{s,u_s}(x)}
}
\right|
&\le
|\beta|\big\|\widehat V_{\rm pe}^\star-\widehat V_{s,u_s}\big\|
\le\frac{|\beta|}{N}.
\end{align*}
Also, the same bound holds for the third term. 
Consequently,
\[
\begin{aligned}
\left|
\log
\frac{
\sum_xP(x|s,a)e^{-\beta\widehat V_{\rm pe}^\star(x)}
}{
\sum_x\widehat P(x|s,a)e^{-\beta\widehat V_{\rm pe}^\star(x)}
}
\right|
&\le
\left|
\log
\frac{
\sum_xP(x|s,a)e^{-\beta\widehat V_{s,u_s}(x)}
}{
\sum_x\widehat P(x|s,a)e^{-\beta\widehat V_{s,u_s}(x)}
}
\right|
+
\frac{2|\beta|}{N}.
\end{aligned}
\]

Using
\eqref{eq:loo-aux-bound}, we obtain
\[
\begin{aligned}
&\frac{\gamma}{|\beta|}
\left|
\log
\frac{
\sum_x\widehat P(x|s,a)e^{-\beta\widehat V_{\rm pe}^\star(x)}
}{
\sum_xP(x|s,a)e^{-\beta\widehat V_{\rm pe}^\star(x)}
}
\right| \le
\frac{4\sqrt{2}}{|\beta|}
\Bigg[
\sqrt{
\frac{
\left[
e^{\frac{|\beta|}{1-\gamma}}
+
e^{-\frac{|\beta|}{1-\gamma}}
-2
\right]
\ell
}{
N(s,a)
}
}
+
\frac{
\left(
e^{\frac{|\beta|}{1-\gamma}}-1
\right)
\ell
}{
N(s,a)
}
\Bigg]
+
\frac{2}{N}.
\end{aligned}
\]
Now let $v$ be any value function satisfying
$\|v-\widehat V_{\rm pe}^\star\|\le 1/N$.
We relate the bound for $\widehat V_{\rm pe}^\star$ to $v$ in the same way as above. We have
\[
\begin{aligned}
\frac{\gamma}{|\beta|}
\left|
\log\frac{\sum_x \widehat P(x|s,a)e^{-\beta v(x)}}
{\sum_x P(x|s,a)e^{-\beta v(x)}}
\right|
&\le
\frac{\gamma}{|\beta|}
\left|
\log\frac{\sum_x \widehat P(x|s,a)e^{-\beta \widehat V_{\rm pe}^\star(x)}}
{\sum_x P(x|s,a)e^{-\beta \widehat V_{\rm pe}^\star(x)}}
\right|
+\frac{2\gamma}{N}
\\
&\le
\frac{4\sqrt{2}}{|\beta|}
\Bigg[
\sqrt{
\frac{
\left[
 e^{\frac{|\beta|}{1-\gamma}}
+
 e^{-\frac{|\beta|}{1-\gamma}}
-2
\right]\ell
}{N(s,a)}}
+
\frac{
\left(e^{\frac{|\beta|}{1-\gamma}}-1\right)\ell
}{N(s,a)}
\Bigg]
+\frac{4}{N}.
\end{aligned}
\]
Furthermore, it is evident that when $N(s,a)=0$, the left-hand side is at most $1/(1-\gamma)$, and the bound follows. 
\end{proof}

\subsubsection{Proof of Lemma \ref{lem:bernstein}}

\begin{proof}
We consider two cases depending on the sign of $\beta$.

\paragraph{Case 1: $\beta>0$.}
Define
\[
        h_{s,a}
        :=
        \sum_xP(x| s,a)e^{-\beta v(x)},
        \qquad
        \widehat h_{s,a}
        :=
        \sum_x\widehat P(x| s,a)e^{-\beta v(x)}.
\]
Since \(0\le v(x)\le \frac{1}{1-\gamma}\) for all $x\in \cS$,  $e^{-|\beta|/(1-\gamma)} \le e^{-\beta v(x)}\le 1$, so that   
\[
e^{-|\beta|/(1-\gamma)} \le h_{s,a} \le 1.
\]
Let $X_1,\ldots,X_{N(s,a)}$ be the next-state samples collected from
$P(\cdot|s,a)$. Then
\[
        \widehat h_{s,a}
        =
        \frac{1}{N(s,a)}
        \sum_{i=1}^{N(s,a)}e^{-\beta v(X_i)}.
\]
The summands are i.i.d., bounded in \([e^{-|\beta|/(1-\gamma)},1]\), and have mean
\(h_{s,a}\). Applying Bernstein's inequality (Lemma \ref{lem:Bernstein}) gives, with probability at least $1-\alpha$,
\begin{align}
\label{eq:bern-h-deviation}
\left|\widehat h_{s,a}-h_{s,a}\right|
&\le
\sqrt{
\frac{
2\operatorname{Var}_{P(\cdot\mid s,a)}\!\left(e^{-\beta v(X)}\right)
\log(2/\alpha)
}{N(s,a)}}
+
\frac{
2\left(1-e^{-\frac{|\beta|}{1-\gamma}}\right)
\log(2/\alpha)
}{3N(s,a)}.
\end{align}
To control $\operatorname{Var}_{P(\cdot\mid s,a)}$, the Bhatia--Davis inequality (Lemma \ref{lem:Bhatia-Davis}) gives
\[
\operatorname{Var}_{P(\cdot\mid s,a)}\!\left(e^{-\beta v(X)}\right)
\le
(1-h_{s,a})
\left(h_{s,a}-e^{-\frac{|\beta|}{1-\gamma}}\right).
\]
Hence, with probability at least $1-\alpha$,
\begin{align}
\label{eq:bern-h-deviation-bd}
\left|\widehat h_{s,a}-h_{s,a}\right|
&\le
\sqrt{
\frac{
2(1-h_{s,a})
\left(h_{s,a}-e^{-\frac{|\beta|}{1-\gamma}}\right)
\log(2/\alpha)
}{N(s,a)}}
+
\frac{
2\left(1-e^{-\frac{|\beta|}{1-\gamma}}\right)
\log(2/\alpha)
}{3N(s,a)}.
\end{align}
We have
\begin{align}
\label{eq:normalized-variance}
&\max_{x\in[e^{-\frac{|\beta|}{1-\gamma}},1]}
\frac{
(1-x)\left(x-e^{-\frac{|\beta|}{1-\gamma}}\right)
}{x^2}
=
\frac14
\left(
 e^{\frac{|\beta|}{1-\gamma}}
 +e^{-\frac{|\beta|}{1-\gamma}}
 -2
\right).
\end{align}
In fact, let $m=e^{-|\beta|/(1-\gamma)}$ and define 
$g(x)=(1-x)(x-m)/x^2$. $g$ is maximized at $x=2m/(1+m)$: 
\[
g\!\left(\frac{2m}{1+m}\right)
=
\frac{(1-m)^2}{4m}
=
\frac{1}{4}\left(
e^{|\beta|/(1-\gamma)}+e^{-|\beta|/(1-\gamma)}-2
\right).
\]
Moreover,
\begin{equation}
\label{eq:normalized-range}
\frac{
1-e^{-\frac{|\beta|}{1-\gamma}}
}{h_{s,a}}
\le
e^{\frac{|\beta|}{1-\gamma}}-1.
\end{equation}
First suppose that the right-hand side of \eqref{eq:bern-h-deviation-bd} is at most $h_{s,a}/2$. Then
$|\widehat h_{s,a}-h_{s,a}|
        \le
        h_{s,a}/2$. 
Applying Lemma \ref{lem:log-inequality} gives
\[
        \left|
        \log\widehat h_{s,a}-\log h_{s,a}
        \right|
        \le
        \frac{2|\widehat h_{s,a}-h_{s,a}|}{h_{s,a}}.
\]
Combining this with \eqref{eq:bern-h-deviation-bd}, \eqref{eq:normalized-variance}, and \eqref{eq:normalized-range} gives
\begin{align}
\label{eq:bern-log-small-deviation}
\left|
\log\widehat h_{s,a}-\log h_{s,a}
\right|
&\le
\sqrt{
\frac{
2\left[
 e^{\frac{|\beta|}{1-\gamma}}
 +e^{-\frac{|\beta|}{1-\gamma}}
 -2
\right]
\log(2/\alpha)
}{N(s,a)}}
+
\frac{
4\left(e^{\frac{|\beta|}{1-\gamma}}-1\right)
\log(2/\alpha)
}{3N(s,a)}.
\end{align}
In particular, this is bounded by
\begin{align}
\label{eq:bern-log-target}
4\sqrt{2}
\left[
\sqrt{
\frac{
\left[
 e^{\frac{|\beta|}{1-\gamma}}
 +e^{-\frac{|\beta|}{1-\gamma}}
 -2
\right]
\log(2/\alpha)
}{N(s,a)}}
+
\frac{
\left(e^{\frac{|\beta|}{1-\gamma}}-1\right)
\log(2/\alpha)
}{N(s,a)}
\right].
\end{align}
It remains to consider the case where the right-hand side of \eqref{eq:bern-h-deviation-bd} is larger than $h_{s,a}/2$.
If $\widehat h_{s,a}\ge h_{s,a}$, then $\log x\le x-1$ gives
\begin{align*}
\left|
\log\widehat h_{s,a}-\log h_{s,a}
\right|
&=
\log\frac{\widehat h_{s,a}}{h_{s,a}}
\\
&\le
\frac{\widehat h_{s,a}-h_{s,a}}{h_{s,a}}
\\
&\le
\frac{1}{\sqrt{2}}
\sqrt{
\frac{
\left[
 e^{\frac{|\beta|}{1-\gamma}}
 +e^{-\frac{|\beta|}{1-\gamma}}
 -2
\right]
\log(2/\alpha)
}{N(s,a)}}
+
\frac{2}{3}
\frac{
\left(e^{\frac{|\beta|}{1-\gamma}}-1\right)
\log(2/\alpha)
}{N(s,a)},
\end{align*}
which is again bounded by \eqref{eq:bern-log-target}.
Now suppose that $\widehat h_{s,a}<h_{s,a}$. Since $\widehat h_{s,a} \ge e^{-\frac{|\beta|}{1-\gamma}}$,
\begin{align*}
\left|
\log\widehat h_{s,a}-\log h_{s,a}
\right|
&=
\log\frac{h_{s,a}}{\widehat h_{s,a}}
\le
\log\left(
 h_{s,a}e^{\frac{|\beta|}{1-\gamma}}
\right).
\end{align*}
Since the right-hand side of \eqref{eq:bern-h-deviation-bd} is larger than $h_{s,a}/2$, at least one of its two terms is larger than $h_{s,a}/4$.
If
\[
\sqrt{
\frac{
2(1-h_{s,a})
\left(h_{s,a}-e^{-\frac{|\beta|}{1-\gamma}}\right)
\log(2/\alpha)
}{
N(s,a)
}}
>
\frac{h_{s,a}}{4},
\]
then, 
\[
\frac{\log(2/\alpha)}{N(s,a)}
>
\frac{h_{s,a}^2}
{32(1-h_{s,a})
\left(h_{s,a}-e^{-\frac{|\beta|}{1-\gamma}}\right)}.
\]
so that 
\[
\begin{aligned}
&
\left[
e^{\frac{|\beta|}{1-\gamma}}
+
e^{-\frac{|\beta|}{1-\gamma}}
-2
\right]
\frac{\log(2/\alpha)}{N(s,a)}
>
\frac{
e^{\frac{|\beta|}{1-\gamma}}
\left(
1-e^{-\frac{|\beta|}{1-\gamma}}
\right)^2
h_{s,a}^2
}{
32(1-h_{s,a})
\left(
h_{s,a}-e^{-\frac{|\beta|}{1-\gamma}}
\right)
}.
\end{aligned}
\]
Since \(h_{s,a}\le 1\),
\[
h_{s,a}-e^{-\frac{|\beta|}{1-\gamma}}
\le
h_{s,a}
\left(
1-e^{-\frac{|\beta|}{1-\gamma}}
\right),
\]
and thus
\[
\left(
h_{s,a}-e^{-\frac{|\beta|}{1-\gamma}}
\right)^2
\le
h_{s,a}^2
\left(
1-e^{-\frac{|\beta|}{1-\gamma}}
\right)^2.
\]
Using also \(1-h_{s,a}\le 1\), we get
\[
\begin{aligned}
&
\frac{
e^{\frac{|\beta|}{1-\gamma}}
\left(
1-e^{-\frac{|\beta|}{1-\gamma}}
\right)^2
h_{s,a}^2
}{
32(1-h_{s,a})
\left(
h_{s,a}-e^{-\frac{|\beta|}{1-\gamma}}
\right)
}
\ge
\frac{
e^{\frac{|\beta|}{1-\gamma}}
\left(
h_{s,a}-e^{-\frac{|\beta|}{1-\gamma}}
\right)
}{32}
=
\frac{
h_{s,a}e^{\frac{|\beta|}{1-\gamma}}-1
}{32}.
\end{aligned}
\]
Therefore,
\[
\left[
e^{\frac{|\beta|}{1-\gamma}}
+
e^{-\frac{|\beta|}{1-\gamma}}
-2
\right]
\frac{\log(2/\alpha)}{N(s,a)}
>
\frac{
h_{s,a}e^{\frac{|\beta|}{1-\gamma}}-1
}{32}.
\]
For $x\ge1$,
$
        \log x\le\sqrt{x-1}.
$
Therefore,
\begin{align*}
\log\left(
 h_{s,a}e^{\frac{|\beta|}{1-\gamma}}
\right)
&\le
\sqrt{
 h_{s,a}e^{\frac{|\beta|}{1-\gamma}}-1
}
<
4\sqrt{2}
\sqrt{
\left[
 e^{\frac{|\beta|}{1-\gamma}}
 +e^{-\frac{|\beta|}{1-\gamma}}
 -2
\right]
\frac{\log(2/\alpha)}{N(s,a)}
}.
\end{align*}
If instead
\[
\frac{
2\left(1-e^{-\frac{|\beta|}{1-\gamma}}\right)
\log(2/\alpha)
}{3N(s,a)}
>
\frac{h_{s,a}}4,
\]
then
\[
\left(e^{\frac{|\beta|}{1-\gamma}}-1\right)
\frac{\log(2/\alpha)}{N(s,a)}
>
\frac{3}{8}
h_{s,a}e^{\frac{|\beta|}{1-\gamma}}.
\]
Using $\log x\le x$ for $x\ge1$,
\begin{align*}
\log\left(
 h_{s,a}e^{\frac{|\beta|}{1-\gamma}}
\right)
&\le
h_{s,a}e^{\frac{|\beta|}{1-\gamma}}
<
\frac{8}{3}
\left(e^{\frac{|\beta|}{1-\gamma}}-1\right)
\frac{\log(2/\alpha)}{N(s,a)},
\end{align*}
which is also bounded by \eqref{eq:bern-log-target}. Thus, for every $N(s,a)>0$,
\begin{align*}
\left|
\log\widehat h_{s,a}-\log h_{s,a}
\right|
&\le
4\sqrt{2}
\left[
\sqrt{
\frac{
\left[
 e^{\frac{|\beta|}{1-\gamma}}
 +e^{-\frac{|\beta|}{1-\gamma}}
 -2
\right]
\log(2/\alpha)
}{N(s,a)}}
+
\frac{
\left(e^{\frac{|\beta|}{1-\gamma}}-1\right)
\log(2/\alpha)
}{N(s,a)}
\right].
\end{align*}
Multiplying by $1/|\beta|$ proves the claim for $\beta>0$.

\medskip
\noindent
\textbf{Case 2: $\beta<0$.}
Define
\[
        h_{s,a}
        :=
        \sum_xP(x| s,a)
        e^{-|\beta|\left(\frac1{1-\gamma}-v(x)\right)},
        \qquad
        \widehat h_{s,a}
        :=
        \sum_x\widehat P(x| s,a)
        e^{-|\beta|\left(\frac1{1-\gamma}-v(x)\right)}.
\]
Compared to the case of $\beta>0$, the exponents have some shift. However, this does not change the logarithmic ratio because it multiplies both \(h_{s,a}\) and
\(\widehat h_{s,a}\) by the same constant \(e^{-|\beta|/(1-\gamma)}\). 
Moreover,
\[
        e^{-\frac{|\beta|}{1-\gamma}}
        \le
        e^{-|\beta|\left(\frac1{1-\gamma}-v(x)\right)}
        \le 1.
\]
Hence the shifted exponential terms lie in exactly the same interval as in Case 1. The same Bernstein inequality, and Bhatia--Davis variance bound therefore give
\begin{align*}
&\left|
\log
\frac{
\sum_x\widehat P(x| s,a)e^{-\beta v(x)}
}{
\sum_xP(x| s,a)e^{-\beta v(x)}
}
\right|
\le
4\sqrt{2}
\left[
\sqrt{
\frac{
\left[
 e^{\frac{|\beta|}{1-\gamma}}
 +e^{-\frac{|\beta|}{1-\gamma}}
 -2
\right]
\log(2/\alpha)
}{N(s,a)}}
+
\frac{
\left(e^{\frac{|\beta|}{1-\gamma}}-1\right)
\log(2/\alpha)
}{N(s,a)}
\right].
\end{align*}
Multiplying by $\gamma/|\beta|$ proves the claim for $\beta<0$.
Finally, if $N(s,a)=0$, then for any probability distribution $q$ and every $v\in[0,1/(1-\gamma)]^S$,
\[
        0
        \le
        -\frac1\beta
        \log\sum_xq(x)e^{-\beta v(x)}
        \le
        \frac1{1-\gamma}.
\]
Thus the absolute difference between the two entropic transition values is at most $1/(1-\gamma)$, and multiplying by $\gamma\le1$ gives the stated deterministic bound.
\end{proof}

\subsection{Proof of Lemma \ref{lem:V_pessimism}}
\begin{proof}
Because $\widehat V_0=0$, and $\PBoper$ is monotone, the value-iteration sequence is non-decreasing. Hence
\[
\widehat V=\widehat V_k
\le
\widehat V_{k+1}
=
\PBoper\widehat V.
\]

On the event $E$,
\[
\left|\frac{\gamma}{\beta}
\log\frac{
\sum_x\widehat P(x\mid s,\widehat\pi(s))e^{-\beta\widehat V(x)}
}{
\sum_xP(x\mid s,\widehat\pi(s))e^{-\beta\widehat V(x)}
}
\right|
\le b(s,\widehat\pi(s)).
\]
Therefore 
\[
R(s,\widehat\pi(s))
-\frac{\gamma}{\beta}\log\!\left(\sum_x\widehat P(x\mid s,\widehat\pi(s))e^{-\beta\widehat V(x)}\right)
-b(s,\widehat\pi(s))
\le
R(s,\widehat\pi(s))
-\frac{\gamma}{\beta}\log\!\left(\sum_xP(x\mid s,\widehat\pi(s))e^{-\beta\widehat V(x)}\right).
\]
By the definition of the output policy $\widehat\pi$, for every state $s$,
\[
(\PBoper\widehat V)(s)
=
\max\left\{0,
 R(s,\widehat\pi(s))
-\frac{\gamma}{\beta}
\log\!\left(\sum_x\widehat P(x\mid s,\widehat\pi(s))e^{-\beta\widehat V(x)}\right)
-b(s,\widehat\pi(s))
\right\}.
\]
And denoting the true fixed-policy Bellman operator by
\[
(\mathcal T^{\widehat\pi}v)(s)
:=
R(s,\widehat\pi(s))
-\frac{\gamma}{\beta}\log\!\left(\sum_xP(x\mid s,\widehat\pi(s))e^{-\beta v(x)}\right),
\]
we obtain
\[
\widehat V
\le
\PBoper\widehat V
\le
\mathcal T^{\widehat\pi}\widehat V.
\]
The operator $\mathcal T^{\widehat\pi}$ is monotone and a $\gamma$-contraction. Iterating the preceding inequality gives
\[
\widehat V
\le
\mathcal T^{\widehat\pi}\widehat V
\le
(\mathcal T^{\widehat\pi})^2\widehat V
\le\cdots,
\]
and the right-hand side converges to the unique fixed point $V^{\widehat\pi}$. Consequently, $\widehat V(s)\le V^{\widehat\pi}(s)$ for all $s\in \cS$, thus proving the lemma. 
\end{proof}

\subsection{Proof of Lemma \ref{lem:weighted-count-bound}}

\begin{proof}
Define 
$
I=\big\{(s,a):N\db(s,a)\ge 8L\big\}
$
with $L=\log(SA/\delta)$, and introduce the event:
$$
\cB = \big\{N(s,a)\ge N\db(s,a)/2, \,\forall (s,a)\in I\big\}\,.
$$
Applying Chernoff's inequality yields
\[
    \mathbb{P}\left(
N(s,a)<\frac12 N\db(s,a)
\right) \le e^{-N\db(s,a)/8}
\le \frac{\delta}{SA},
\]
so that by a union bound, we have that $\PP(\cB)\ge 1-\delta$. 

The definition of $\Cclipped$ implies that 
\begin{align}
    \label{eq:Cclipped_implication}
\min\left\{d^\star(s,a),\frac1S\right\}
\le \Cclipped\db(s,a), \qquad \forall (s,a). 
\end{align}
In view of this, we  partition  the state-action pairs into
$J=\{(s,a):d^\star(s,a)\le 1/S\}$ and $J^c = (\cS\times \cA)\setminus J$. 

\paragraph{Proof of (i).} Recall that $\cS\setminus\cS^+ =\{s: N(s,\pi^\star(s))=0\}$. Assume that $\cB$ holds. Then every $(s,a)\in I$ satisfies 
$N(s,a)\ge 4L\ge 1$ (since $SA\ge 2$). Hence
$\cS\setminus\cS^+ \subseteq I^c$, 
so that 
$$
\sum_{s\in \cS\setminus\cS^+} d^\star(s) \indOne\{\cB\} \le \sum_{(s,a)\in I^c} d^\star(s,a) \indOne\{\cB\}.
$$
For pairs in $J$, we have that 
$d^\star(s,a)\le \Cclipped\db(s,a)$.
Thus, for $(s,a)\in J\cap I^c$,
\[
d^\star(s,a)
\le \Cclipped\db(s,a)
< \frac{8\Cclipped L}{N}.
\]
Since the support of $d^\star$ contains at most $S$
pairs,
\[
\sum_{(s,a)\in J\cap I^c}d^\star(s,a)
\le \frac{8S\Cclipped L}{N}.
\]
For pairs not in $J$, (\ref{eq:Cclipped_implication}) implies that $\db(s,a)\ge \frac{1}{S\Cclipped}$. Hence, if $J^c\cap I^c$ is non-empty, then
\[
\frac{N}{S\Cclipped}
\le N\db(s,a)<8L,
\]
so that $N<8S\Cclipped L$. Using
$\sum_{s,a}d^\star(s,a)=1$, we obtain
\[
\sum_{(s,a)\in J^c \cap I^c}d^\star(s,a)
\le 1<\frac{8S\Cclipped L}{N}.
\]
The same upper bound holds trivially if
$J_2\cap I^c$ is empty. 

Putting together, 
\[
\sum_{(s,a)\in I^c} d^\star(s,a) \indOne\{\cB\}
\le \frac{16S\Cclipped L}{N}.
\]
Noting that $\cB$ holds with probability at least $1-\delta$ establishes (i). 

\paragraph{Proof of (ii).} 
Assume that $\cB$ holds. For $(s,a)\in J\cap I$, by (\ref{eq:Cclipped_implication}) and definition of $\cB$, we have 
\[
\frac{d^\star(s,a)}{\max\{N(s,a),1\}}
\le \frac{2d^\star(s,a)}{N\db(s,a)}
\le \frac{2\Cclipped}{N}
\le \frac{8\Cclipped L}{N},
\]
where the last inequality uses $L\ge\log 2>1/2$.

For $(s,a)\in J \cap I^c$,
\[
\frac{d^\star(s,a)}{\max\{N(s,a),1\}}
\le d^\star(s,a)
<\frac{8\Cclipped L}{N}.
\]
Since at most $S$ pairs with $d^\star(s)>0$, combining these two bounds yields
\[
\sum_{(s,a)\in J}
\frac{d^\star(s,a)}{\max\{N(s,a),1\}}
\le \frac{8S\Cclipped L}{N}.
\]

For $(s,a)\in J^c\cap I$, we have
$\db(s,a)\ge 1/(S)$. Therefore,
\begin{align*}
\sum_{(s,a)\in J^c\cap I}
\frac{d^\star(s,a)}{\max\{N(s,a),1\}}
&\le
\frac{2}{N}
\sum_{(s,a)\in J^c\cap I}
\frac{d^\star(s,a)}{\db(s,a)}
\\
&\le
\frac{2S\Cclipped}{N}
\sum_{(s,a)\in J^c\cap I}d^\star(s,a)
\\
&\le \frac{2S\Cclipped}{N}
\le \frac{4S\Cclipped L}{N}.
\end{align*}
Finally, for  $(s,a)\in J^c\cap I^c$, $N(s,a)\ge 1$ so that 
\[
\sum_{(s,a)\in J^c\cap I^c}
\frac{d^\star(s,a)}{\max\{N(s,a),1\}}
\le
\sum_{(s,a)\in J^c\cap I^c}d^\star(s,a)
\le \frac{8S\Cclipped L}{N}.
\]
Putting the four cases together, we conclude that on the event $\cB$, (ii) holds. 
\end{proof}

\section{Proof of the Minimax Lower Bound}

\subsection{Preliminary Facts}
In this section we provide the proof of the lower bound along with some preliminary lemmas:

\begin{lemma}[Bernoulli KL bound]\label{lem:bernoulli-kl}
Let \(0<p<p+\alpha<1\). Then
\[
\begin{aligned}
    &\KL\bigl(p+\alpha, p\bigr)
    +
    \KL\bigl(p, p+\alpha\bigr)
    \le
    \frac{\alpha^2}{p(1-p-\alpha)}.
\end{aligned}
\]
\end{lemma}

\begin{proof}
By direct expansion,
\begin{align*}
\KL(p+\alpha, p) &+\KL(p, p+\alpha) \\
&=
(p+\alpha)\log\frac{p+\alpha}{p}
+(1-p-\alpha)\log\frac{1-p-\alpha}{1-p}
+p\log\frac{p}{p+\alpha}
+(1-p)\log\frac{1-p}{1-p-\alpha}
\\
&= \alpha\log\frac{p+\alpha}{p}
+\alpha\log\frac{1-p}{1-p-\alpha}.
\end{align*}
Using $\log(1+x)\leq x$,
\begin{align*}
\KL(p+\alpha, p) +\KL(p, p+\alpha)
&=
\alpha\log\left(1+\frac{\alpha}{p}\right)
+
\alpha\log\left(1+\frac{\alpha}{1-p-\alpha}\right) \\
&\leq \alpha^2
\left(\frac{1}{p} +
\frac{1}{1-p-\alpha}
\right)\\
&=
\frac{\alpha^2(1-\alpha)}
{p(1-p-\alpha)} \\
&\leq
\frac{\alpha^2}
{p(1-p-\alpha)}.
\end{align*}
\end{proof}

\begin{lemma}[Assouad and conversion to a constant-probability event]\label{lem:assouad}
Let \(\{\mathbb P_\theta:\theta\in\{0,1\}^m\}\) be a family of distributions. For \(i\in[m]\), let \(\theta^{(i)}\) be the vector obtained from \(\theta\) by flipping only the \(i\)-th coordinate. Suppose
\[
    \KL(\mathbb P_\theta, \mathbb P_{\theta^{(i)}})\le \frac18, 
    \qquad\forall \theta\in\{0,1\}^m,\ \forall i\in[m].
\]
Then every estimator \(\widehat\theta\in\{0,1\}^m\) satisfies
\[
    \sup_{\theta\in\{0,1\}^m}
    \mathbb P_\theta\!\left(
        \Ham(\widehat\theta,\theta)\ge \frac{m}{8}
    \right)
    \ge \frac18.
\]
\end{lemma}

\begin{proof}
The Kullback-Leibler version of Assouad's Lemma
\cite[Theorem~2.12(iv)]{tsybakov2009introduction}, gives
\begin{align*}
\inf_{\widehat\theta}
\sup_{\theta\in\{0,1\}^m}
\mathbb E_\theta
\left[
\Ham(\widehat\theta,\theta)
\right]
&\ge
\frac m2
\max\left\{
\frac12\exp\left(-\frac18\right),
1-\sqrt{\frac{1}{16}}
\right\}
\nonumber
=
\frac{3m}{8}.
\end{align*}
Therefore, for any estimator $\widehat\theta$, there exists
$\theta\in\{0,1\}^m$ such that
\begin{align}
\mathbb E_\theta
\left[
\Ham(\widehat\theta,\theta)
\right]
\ge
\frac{3m}{8}.
\label{eq:assouad-expected-hamming}
\end{align}

On the other hand, since
$0\le\Ham(\widehat\theta,\theta)\le m$,
\begin{align}
\mathbb E_\theta
\left[
\Ham(\widehat\theta,\theta)
\right]
&=
\mathbb E_\theta
\left[
\Ham(\widehat\theta,\theta)
\indOne\left\{
\Ham(\widehat\theta,\theta)<\frac m8
\right\}
\right]
\nonumber
+
\mathbb E_\theta
\left[
\Ham(\widehat\theta,\theta)
\indOne\left\{
\Ham(\widehat\theta,\theta)\ge\frac m8
\right\}
\right]
\nonumber\\
&\le
\frac m8
+
m\,
\mathbb P_\theta
\left(
\Ham(\widehat\theta,\theta)\ge\frac m8
\right).
\label{eq:assouad-expectation-to-probability}
\end{align}
Combining
\eqref{eq:assouad-expected-hamming}
and
\eqref{eq:assouad-expectation-to-probability}
gives
\[
\mathbb P_\theta
\left(
\Ham(\widehat\theta,\theta)\ge\frac m8
\right)
\ge
\frac14.
\]
Consequently, for any estimator $\widehat\theta$,
\[
\sup_{\theta\in\{0,1\}^m}
\mathbb P_\theta
\left(
\Ham(\widehat\theta,\theta)\ge\frac m8
\right)
\ge
\frac14
\ge
\frac18,
\]
which proves the claim.
\end{proof}

\subsection{Proof of Theorem \ref{thm:LB_ERM}}

\begin{proof}
\medskip\noindent\textbf{Step 1: construction of the hard MDPs.}
Let $m:=\lceil S/4\rceil$.
For each vector \(\theta\in\{0,1\}^m\), define the discounted MDP
$M_\theta=(\cS,\cA,P_\theta,r,\gamma,\nu)$ as follows. Throughout the proof, we use the shorthand
$V_\theta^\pi:=V_{M_\theta}^\pi$, $V_\theta^\star:=V_{M_\theta}^\star$, and $Q_\theta^\star:=Q_{M_\theta}^\star$. The state and action spaces are
\[
    \cS=\{s_1,\ldots,s_m,s_G,s_B,s_{m+3},\ldots,s_S\},
    \qquad
    \cA=\{0,1,\ldots,A-1\}.
\]
The states \(s_1,\ldots,s_m\) are decision states, \(s_G\) is the good absorbing state, \(s_B\) is the bad absorbing state, and \(s_{m+3},\ldots,s_S\) are dummy states. At each decision state only actions \(0\) and \(1\) are informative. For $s_i$, \(\theta_i\in \{0,1\}\) specifies which of these two actions is better.

For parameters \(p\) and \(\alpha\) chosen below, the transition kernel of \(M_\theta\) is
\begin{equation}
P_\theta(s'\mid s,a)=
\begin{cases}
    (p+\alpha)\indOne\{s'=s_G\}+(1-p-\alpha)\indOne\{s'=s_B\},
        & s=s_i,\ a=\theta_i,\ i\in[m],\\
    p\indOne\{s'=s_G\}+(1-p)\indOne\{s'=s_B\},
        & s=s_i,\ a=1-\theta_i,\ i\in[m],\\
    \indOne\{s'=s_B\},
        & s=s_i,\ a\in\cA\setminus\{0,1\},\ i\in[m],\\
    \indOne\{s'=s_G\},
        & s=s_G,\ a\in\cA,\\
    \indOne\{s'=s_B\},
        & s=s_B,\ a\in\cA,\\
    \indOne\{s'=s\},
        & s\in\{s_{m+3},\ldots,s_S\},\ a\in\cA.
\end{cases}
\label{eq:transition-kernel-hard-family}
\end{equation}
Equivalently, for each decision state \(s_i\), the only two informative rows of the transition matrix are
\[
\begin{array}{c|cc}
\text{state-action pair} & s_G & s_B  \\
\hline
(s_i,\theta_i)     & p+\alpha & 1-p-\alpha \\
(s_i,1-\theta_i)   & p        & 1-p
\end{array}
\]
The reward function is the same for every \(M_\theta\), and is given by $R(s,a) = \indOne\{s=s_G\}$ for all $a\in \cA$. 
Thus, for every \(M_\theta\),
\[
    V^\star_\theta(s_G)=\frac{1}{1-\gamma},
    \qquad
    V^\star_{\theta}(s_B)=0.
\]
The initial (test) distribution is chosen as the uniform distribution over the decision states:
\[
    \nu(s_i)=\frac1m,\qquad i=1,\ldots,m.
\]
It remains to choose $p$ and $\alpha$ so that action \(\theta_i\) is better than action \(1-\theta_i\) by a visible value gap, while the two transition rows remain statistically hard to distinguish from offline data. For every decision state-action pair \((s_i,a)\), the next state is either \(s_G\) or \(s_B\), and the immediate reward is $0$. Hence,
\begin{equation}
    Q^\star_{\theta}(s_i,a)
    =
    -\frac{\gamma}{\beta}
    \log\left(
        P_\theta(s_G\mid s_i,a)\exp\!\left(-\frac{\beta}{1-\gamma}\right)
        +1-P_\theta(s_G\mid s_i,a)
    \right).
    \label{eq:explicit-Q-both-signs}
\end{equation}
Setting 
\begin{equation}
\label{eq:new-p-alpha}
    p:=
    \frac{1}{1+\exp\!\left(-\frac{\beta}{1-\gamma}\right)},
    \qquad
    \alpha:=
    \frac{128|\beta|\varepsilon}
    {\gamma\left(
    \exp\!\left(\frac{|\beta|}{1-\gamma}\right)
    -
    \exp\!\left(-\frac{|\beta|}{1-\gamma}\right)
    \right)},
\end{equation}
by the assumption on \(\varepsilon\), we get 
\begin{equation}
\label{ineq:alpha-small-new}
    \alpha
    \le
    \frac{1}{2}\left(1+\exp\!\left(\frac{|\beta|}{1-\gamma}\right)\right)^{-1}
    =     \frac12\min\{p,1-p\}.
\tag{Q1}
\end{equation}
Therefore \(0<p<p+\alpha<1\), and the kernel in \eqref{eq:transition-kernel-hard-family} is well-defined.

\begin{lemma}\label{lem:Q_diff_LB}
For every decision state $s_i$:
\begin{equation}
\label{eq:new-Q-gap}
    Q^\star_{\theta}(s_i,\theta_i)
    -
    Q^\star_{\theta}(s_i,1-\theta_i)
    \ge32\varepsilon.
\tag{Q3}
\end{equation}
\end{lemma}

\medskip\noindent\textbf{Step 2: offline distribution and concentrability.}
The offline state-action sampling distribution is the same for all \(M_\theta\). Define
\begin{equation}
    \db(s_i,a)=\frac{1-\gamma}{mC},
    \qquad i=1,\ldots,m,
    \qquad a\in\{0,1\}.
    \label{eq:mu-decision}
\end{equation}
Fix one absorbing action \(a_{\mathrm{abs}}\in\cA\). Assign behavior mass at least \(1/(SC)\) to \((s_G,a_{\mathrm{abs}})\) and \((s_B,a_{\mathrm{abs}})\), and put the remaining probability mass on dummy state-action pairs. This is feasible because
\[
    \sum_{i=1}^m\sum_{a\in\{0,1\}}\frac{1-\gamma}{mC}
    +\frac{2}{SC}
    =
    \frac{2(1-\gamma)}{C}+\frac{2}{SC}
    \le
    \frac{1}{2C}+\frac{1}{4C}<1.
\]

For each $\theta$, choose an optimal deterministic policy \(\pi_\theta^\star\) with
\[
    \pi_\theta^\star(s_i)=\theta_i,\qquad i\in[m],
    \qquad
    \pi_\theta^\star(s_G)=\pi_\theta^\star(s_B)=a_{\mathrm{abs}},
\]
and let $d_\theta^\star$ denote its discounted occupancy distribution in $M_\theta$ from the initial distribution $\nu$. The state-action pair \((s_i,\theta_i)\) is visited only at time zero. Therefore,
\[
    d_\theta^\star(s_i,\theta_i)
    =
    \frac{1-\gamma}{m}.
\]
Because \(m=\lceil S/4\rceil\) and \(\gamma\ge3/4\), we have \(m/S\ge1/4\ge1-\gamma\). Hence
$\frac{1-\gamma}{m}\le \frac1S$, and for every $i\in[m]$, 
\[
    \frac{\min\{d_\theta^\star(s_i,\theta_i),1/S\}}
    {\db(s_i,\theta_i)}
    =
    \frac{(1-\gamma)/m}{(1-\gamma)/(mC)}
    =C.
\]
For absorbing state-action pairs, the clipped numerator is at most \(1/S\), and the behavior probability is at least \(1/(SC)\), so the corresponding ratio is at most \(C\). Suboptimal decision-state actions and dummy pairs have zero optimal occupancy. Therefore,
\[
\Cclipped(M_\theta,\db,\nu)=C,
    \qquad \forall\theta\in\{0,1\}^m,
\]
as required by Theorem~\ref{thm:LB_ERM}.

\medskip\noindent\textbf{Step 3: KL divergence between \(M_\theta\) and \(M_{\theta^{(i)}}\).}
Let \(\mathbb P_\theta\) be the distribution of the offline dataset under \(M_\theta\). Each data point is generated by first sampling a state-action pair from the same distribution \(\db\), and then sampling the next state from the transition row of the corresponding MDP. Since the sampling distribution \(\db\) is independent of \(\theta\), the chain rule for KL divergence gives, for any \(\theta\in\{0,1\}^m\) and \(i\in[m]\), where \(\theta^{(i)}\) is obtained from \(\theta\) by flipping only its \(i\)-th coordinate, 
\[
\begin{aligned}
    \KL(\mathbb P_\theta, \mathbb P_{\theta^{(i)}})
    &=
    N\sum_{s,a}\db(s,a)
    \KL\!\left(
        P_\theta(\cdot\mid s,a),
        P_{\theta^{(i)}}(\cdot\mid s,a)
    \right).
\end{aligned}
\]
The two MDPs are identical on every transition row except the two rows \((s_i,0)\) and \((s_i,1)\). On one of these two rows the next-state distribution changes from \(\Bern(p+\alpha)\) to \(\Bern(p)\), and on the other row it changes from \(\Bern(p)\) to \(\Bern(p+\alpha)\). Therefore, using \(\db(s_i,0)=\db(s_i,1)=(1-\gamma)/(mC)\),
\[
\begin{aligned}
    \KL(\mathbb P_\theta, \mathbb P_{\theta^{(i)}})
    &=
    N\frac{1-\gamma}{mC}
    \Bigl[
        \KL(p+\alpha, p)
        +
        \KL(p, p+\alpha)
    \Bigr] \\
    &\le
    N\frac{1-\gamma}{mC}
    \frac{\alpha^2}{p(1-p-\alpha)} .
\end{aligned}
\]
The last inequality is exactly the Bernoulli KL bound of Lemma~\ref{lem:bernoulli-kl}. By \eqref{ineq:alpha-small-new},
$1-p-\alpha\ge\frac{1-p}{2}$, and hence
\[
    p(1-p-\alpha)\ge\frac12p(1-p).
\]
Consequently,
\begin{equation}
\label{eq:new-KL-before-identity}
    \KL(\mathbb P_\theta, \mathbb P_{\theta^{(i)}})
    \le
    2N\frac{1-\gamma}{mC}
    \frac{\alpha^2}{p(1-p)}.
\end{equation}
For the choices of \(p\) and \(\alpha\) in \eqref{eq:new-p-alpha}, a direct calculation gives
\[
\begin{aligned}
    \frac{\alpha^2}{p(1-p)}
    &\le
    2^{14}
    \frac{|\beta|^2\varepsilon^2}{\gamma^2}
    \frac{1}{
    \exp\!\left(\frac{|\beta|}{1-\gamma}\right)
    +
    \exp\!\left(-\frac{|\beta|}{1-\gamma}\right)
    -2
    }.
\end{aligned}
\]
Substituting this into \eqref{eq:new-KL-before-identity} yields
\begin{equation}
\label{eq:new-neighbor-KL}
\begin{aligned}
    \KL(\mathbb P_\theta, \mathbb P_{\theta^{(i)}})
    &\le
    2^{15}
    N\frac{1-\gamma}{mC}
    \frac{|\beta|^2\varepsilon^2}{\gamma^2}
    \frac{1}{
    \exp\!\left(\frac{|\beta|}{1-\gamma}\right)
    +
    \exp\!\left(-\frac{|\beta|}{1-\gamma}\right)
    -2
    }.
\end{aligned}
\end{equation}
Therefore, if
\[
    N\le
    \frac{1}{2^{18}}
    \frac{\gamma^2mC}
    {(1-\gamma)|\beta|^2\varepsilon^2}
    \left[
    \exp\!\left(\frac{|\beta|}{1-\gamma}\right)
    +
    \exp\!\left(-\frac{|\beta|}{1-\gamma}\right)
    -2
    \right],
\]
then 
\[
\KL(\mathbb P_\theta, \mathbb P_{\theta^{(i)}})\le \frac{1}{8},
    \qquad
    \forall\theta\in\{0,1\}^m,\ \forall i\in[m]
\]
This is exactly the hypothesis of Lemma~\ref{lem:assouad}.

\paragraph{Step 4: decoding the hidden actions from the output policy.}
Given the output policy \(\widehat\pi\), define
\[
    \widehat\theta_i\in\argmax_{a\in\{0,1\}}\widehat\pi(a\mid s_i),
    \qquad i=1,\ldots,m.
\]
If \(\widehat\pi\) is deterministic, this is the action selected at \(s_i\). If \(\widehat\pi\) is randomized and \(\widehat\theta_i\ne\theta_i\), then \(\widehat\pi(\theta_i\mid s_i)\le1/2\).
By Lemma~\ref{lem:assouad}, there exists \(\theta\in\{0,1\}^m\) such that
\begin{equation}
    \mathbb P_\theta\!\left(
        \Ham(\widehat\theta,\theta)\ge\frac{m}{8}
    \right)
    \ge \frac18.
    \label{eq:hamming-event}
\end{equation}
We show that this event implies policy suboptimality. Suppose \(\widehat\theta_i\ne\theta_i\), and set
$p_i:=\widehat\pi(\theta_i\mid s_i)$. 
Then \(p_i\le 1/2\), because the decoder chooses the more likely action among \(0\) and \(1\).
At state \(s_i\), the action \(\theta_i\) goes to \(s_G\) with probability \(p+\alpha\), the action \(1-\theta_i\) goes to \(s_G\) with probability \(p\), and every other action goes to \(s_B\). Therefore, under \(\widehat\pi\), the probability of moving to \(s_G\) is at most
\[
    p_i(p+\alpha)+(1-p_i)p
    =p+p_i\alpha
    \le p+\frac{\alpha}{2}.
\]
Since \(F\) is increasing,
\[
\begin{aligned}
    V^\star_{\theta}(s_i)-V^{\widehat\pi}_{\theta}(s_i)
    &\ge
    F(p+\alpha)-F\left(p+\frac{\alpha}{2}\right).
\end{aligned}
\]
For \(\beta>0\), \(F'\) is non-decreasing, so
\[
    F(p+\alpha)-F\left(p+\frac{\alpha}{2}\right)
    \ge
    \frac{\alpha}{2}F'(p)
    =32\varepsilon.
\]
For \(\beta<0\), the argument used in Step 1 gives \(F'(q)\ge F'(p)/2\) for every \(q\in[p,p+\alpha]\). Hence
\[
    F(p+\alpha)-F\left(p+\frac{\alpha}{2}\right)
    \ge
    \frac{\alpha}{4}F'(p)
    =16\varepsilon.
\]
Thus, for both signs of \(\beta\),
\[
    V^\star_{\theta}(s_i)-V^{\widehat\pi}_{\theta}(s_i)
    \ge16\varepsilon.
\]
Therefore, on the event \(\Ham(\widehat\theta,\theta)\ge m/8\),
\[
    V^\star_{\theta}(\nu)-V^{\widehat\pi}_{\theta}(\nu)
    \ge
    \frac1m\cdot\frac{m}{8}\cdot 16\varepsilon
    =2\varepsilon>\varepsilon.
\]
Combining this implication with \eqref{eq:hamming-event} gives
\[
    \sup_\theta \mathbb P_\theta \!\left(
        V^\star_\theta(\nu)-V^{\widehat\pi}_\theta(\nu)>\varepsilon
    \right)
    \ge \frac18.
\]
Since \(m=\lceil S/4\rceil\ge S/4\),
\[
    N\le
    \frac{1}{2^{20}}
    \frac{\gamma^2SC}
    {(1-\gamma)|\beta|^2\varepsilon^2}
    \left[
    \exp\!\left(\frac{|\beta|}{1-\gamma}\right)
    +
    \exp\!\left(-\frac{|\beta|}{1-\gamma}\right)
    -2
    \right]
\]
is sufficient for the KL condition of Lemma~\ref{lem:assouad}. This proves the theorem. 
\end{proof}

\subsection{Technical Lemmas for Lower Bound}

\begin{proof} (of Lemma \ref{lem:Q_diff_LB}). Let
\[
    F(q):=
    -\frac{\gamma}{\beta}
    \log\left(
        q\exp\!\left(-\frac{\beta}{1-\gamma}\right)+1-q
    \right).
\]
Then \(F\) is increasing for every \(\beta\neq0\), and
\begin{equation}
\label{eq:F-derivative}
    F'(q)
    =
    \frac{\gamma\left(1-\exp\!\left(-\frac{\beta}{1-\gamma}\right)\right)}
    {\beta\left(q\exp\!\left(-\frac{\beta}{1-\gamma}\right)+1-q\right)}.
\end{equation}
At the choice of \(p\) in \eqref{eq:new-p-alpha},
\begin{equation}
\label{eq:F-derivative-at-p}
    F'(p)
    =
    \frac{\gamma}{2|\beta|}
    \left(
        \exp\!\left(\frac{|\beta|}{1-\gamma}\right)
        -
        \exp\!\left(-\frac{|\beta|}{1-\gamma}\right)
    \right),
\end{equation}
so that
\begin{equation}
\label{eq:alpha-Fprime}
    \alpha F'(p)=64\varepsilon.
\tag{Q2}
\end{equation}
For \(\beta>0\), \(F\) is convex and therefore \(F'\) is nondecreasing. Hence
\[
\begin{aligned}
    F(p+\alpha)-F(p)
    &=
    \int_p^{p+\alpha} F'(q)\,\mathrm dq 
    \ge
    \int_p^{p+\alpha} F'(p)\,\mathrm dq 
    =
    \alpha F'(p)
    =64\varepsilon.
\end{aligned}
\]
For \(\beta<0\), \(F\) is concave and \(F'\) is non-increasing. By \eqref{ineq:alpha-small-new}, \(\alpha\le p/2\). For every \(q\in[p,p+\alpha]\), the denominator in \eqref{eq:F-derivative} is at most twice its value at \(p\), and therefore
\[
    F'(q)\ge \frac12F'(p).
\]
Consequently,
\[
\begin{aligned}
    F(p+\alpha)-F(p)
    &=
    \int_p^{p+\alpha} F'(q)\,\mathrm dq \ge
    \int_p^{p+\alpha} \frac12F'(p)\,\mathrm dq =
    \frac{\alpha}{2}F'(p)
    =32\varepsilon.
\end{aligned}
\]
Thus, for both cases of \(\beta\),
\begin{equation*}
    Q^\star_{\theta}(s_i,\theta_i)
    -
    Q^\star_{\theta}(s_i,1-\theta_i)
    =F(p+\alpha)-F(p)
    \ge32\varepsilon.
\end{equation*}
\end{proof}

\section{Supporting Lemmas}

\begin{lemma}
\label{lem:log-inequality}
Consider $x,y>0$. If
$|y-x|\le \frac{1}{2} x$, then
\[
|\log y-\log x|
\le
\frac{2|y-x|}{x}.
\]
\end{lemma}

\begin{proof}
First, we consider $x>y$.  By the mean value theorem applied to \(x\mapsto \log x\), there exists some \(\xi\in[y,x]\) such that
\[
\log y-\log x=
\frac{1}{\xi}(y- x).
\]
By the assumption, $y\ge\frac12 x$, so that any \(\xi\in[y,x]\) satisfies
$\xi\ge \frac{1}{2}x$, 
thus giving 
\[
|\log y-\log x|
=
\frac{|y-x|}{\xi}
\le
\frac{2|y-x|}{x}.
\]
If $x<y$, then $\xi\ge x$ will hold trivially, and the lemma follows. 
\end{proof}

\begin{lemma}[{Bernstein's Inequality}] \label{lem:Bernstein}
Let $(X_1,\ldots,X_n)$ be independent, zero-mean random variables such that
$|X_i|\le b$ almost surely, and define $V:=\sum_{i=1}^n \mathbb{E}[X_i^2]$.
Then, for any $\delta\in(0,1)$, with probability at least \(1-\delta\),
$$
\sum_{i=1}^n X_i
\le
\sqrt{2V\log(1/\delta)} + \frac{b}{3}\log(1/\delta).
$$
\end{lemma}

\begin{lemma}[{The Bhatia--Davis Inequality \citep{bhatia2000better}}] \label{lem:Bhatia-Davis}
Let $X$ be a random variable such that $m \le X \le M$ almost surely, and let $\mu = \mathbb{E}[X]$. Then
\[
\operatorname{Var}(X) \le (M-\mu)(\mu-m).
\]
In particular,
\[
\operatorname{Var}(X) \le \frac{(M-m)^2}{4}.
\]
\end{lemma}

\end{document}